%% file: arxiv.tex
\documentclass{article} 
\usepackage{iclr2026_conference,times}

\input{math_commands.tex}

\usepackage{hyperref}
\usepackage{url}
\usepackage{hyperref}       
\usepackage{url}            
\usepackage{booktabs}       
\usepackage{amsfonts}       
\usepackage{amssymb}        
\usepackage{nicefrac}      
\usepackage[table,dvipsnames]{xcolor}         
\usepackage{tabularx}       
\usepackage{multirow}       
\usepackage{array}          
\usepackage{makecell}       
\usepackage{adjustbox}      
\usepackage{threeparttable} 
\usepackage{amsmath}        
\usepackage{caption}        
\usepackage{longtable}
\usepackage{wrapfig}
\usepackage{siunitx}
\usepackage{pgf}            
\usepackage{graphicx}       
\usepackage{import}         

\usepackage[capitalize]{cleveref}
\usepackage{comment}
\usepackage{csquotes}
\usepackage{wrapfig}
\usepackage{mathtools}
\usepackage{subcaption}
\usepackage{etoolbox}
\makeatletter
\patchcmd{\@maketitle}%
  {Published as a conference paper at ICLR 2027}%
  {Preprint.}%
  {}%
  {}
\makeatother

\newcommand{\mypara}[1]{\textbf{\textit{#1}.}}

\title{Detecting Fine-Grained Differences between Image Datasets}

\author{
\noindent
\begin{tabular*}{\linewidth}{@{\extracolsep{\fill}} llll @{}}
Leander Girrbach\textsuperscript{\normalfont 1,2} & 
Alexander von Recum\textsuperscript{\normalfont 3} &
Kirill Bykov\textsuperscript{\normalfont 1,2} &
Zeynep Akata\textsuperscript{\normalfont 1,2} \\[0.6em]
\multicolumn{4}{@{}l@{}}{\normalfont \textsuperscript{1}Technical University of Munich, Munich Center for Machine Learning (MCML)} \\
\multicolumn{4}{@{}l@{}}{\normalfont \textsuperscript{2}Helmholtz Munich \qquad \textsuperscript{3}Ludwig Maximilian University of Munich}
\end{tabular*}
}

\iclrfinalcopy
\begin{document}

\maketitle

\begin{abstract}
Detecting whether and in particular \textit{how} two data distributions differ is a fundamental statistical problem. For example, we might be interested in which concepts (expressed in natural language) differ between two classes, the outputs of different generative models, or the same model with different prompt specifications. When working with large and complex image datasets, we intuitively think that first selecting concepts from a broad list using fine-grained patch embeddings, and then optimizing the concepts to better fit the data is an effective workflow. However, prior work instead uses MLLMs to propose concepts and image-level embeddings to select them. Therefore, we systematically compare these techniques and find that selecting concepts from broad lists can replace MLLM-proposals at obviously lower cost. MLLM inference can then be spent on optimizing selected concepts instead to achieve substantial improvements. Additionally, by designing a new benchmark that focuses on low-prevalence and multi-concept distinctions, we find that selecting nonredundant concepts instead of only the most distinguishing concepts matters, and we introduce techniques for how to effectively do this. Finally, we show how our methods address important problems like detecting spurious correlations and biases. Overall, our work presents a fundamental perspective shift on how we analyze interpretable distributional differences between datasets.
\end{abstract}

\section{Introduction}
\label{sec:introduction}

Many methods can tell us that two datasets differ, but comparatively few can explain what makes them different in human-interpretable terms. For example, classification separates class-conditional distributions, and distribution-shift detection compares training and deployment distributions. However, most methods tell us only whether or how much two distributions differ. To make this understandable to humans and amenable to analyses, we also need to find out \textit{what} makes them different. \textit{Dataset diffing} addresses this by searching for concepts (expressed in natural language) whose prevalence differs between two datasets. For images, which are the focus of this paper, this requires jointly solving three challenges: perceiving evidence in images, aggregating it across the entire data, and expressing the resulting distinction in language.

Beyond being an interesting problem, dataset diffing has many important and practical applications. For example, dataset diffing methods are used for dataset organization \citep{liu2026organizing}, temporal trend analysis in city scenes \citep{deng2025visual} and art \citep{olshausen2010statistics,ginosar2015century}, understanding differences between generative models \citep{jiang2025interpretable,dunlap2025discovering}, and understanding classifier failure modes \citep{eyuboglu2022domino,dunlap2024describing}. Furthermore, dataset diffing can be extended to detecting spurious correlations \citep{kim2024discovering,choi2025conceptscope} and biases \citep{dinca2024openbias}, which are the focus of our applications.

Our approach to dataset diffing is as follows: We start with a broad and general wordlist that captures the fundamental concepts that exist in language. We then use this list to efficiently select the rough concepts whose prevalence differs between the compared datasets. To detect their prevalence, we use patch embeddings, which function as a zero-shot fine-grained concept detector. Image-level embeddings cannot achieve this, because they often submerge details. Finally, we optionally optimize the concepts using an MLLM that receives sampled images and selected concepts and outputs improved concept candidates. In comparison to existing literature, this is a fundamental shift: We don't derive concepts directly from the data, e.g., by repeatedly proposing them with an MLLM, but triangulate the data with a general list that can test a large number of concepts efficiently, and then optimize them. Prior work proposes a number of concepts and then ranks them, and we first select concepts and then optimize them. In addition, we focus on complex datasets that are distinguished by multiple concepts with individually low prevalence, which requires nonredundant concept selection, because many semantically related concepts tend to score similarly while not adding information and instead suppressing other differences. This focus is also novel in comparison to existing benchmarks that focus on individual distinguishing concepts that are represented in all images.

In summary, our contributions are as follows: 
(1) We fill an important gap in the literature by systematically demonstrating the advantages and opportunities of fine-grained, patch-level image representations for dataset diffing;
(2) We show that (expensive) MLLM proposals can be replaced by taking concepts from a broad, static wordlist, and we propose to spend the MLLM budget on post-selection concept optimization instead.
(3) We show that standard benchmarks (VisDiffBench and Metashift) are saturated and propose a more challenging setting where ground truth concepts are sparser and noisier. We show that in this setting, redundancy-aware concept selection and coverage-seeking MLLM optimization are necessary for strong performance, but no method saturates the benchmark; 
(4) We demonstrate through two relevant applications (finding spurious correlations and demographic biases in text-to-image models) that patch-level embeddings and dataset diffing together enable insights at scale that are hard to achieve with image-level embeddings alone. Hence, this paper is a relevant contribution to the community with high practical value.

\section{Related Work}
\label{sec:related}

\mypara{Dataset Diffing in Image Datasets}
Existing approaches for image dataset diffing can be categorized into two main strategies. The first approach uses external models to convert each image into a bag-of-visual-words representation. This process extracts objects and attributes in an image through captioning \citep{kim2024discovering,liu2026organizing}, image tagging \citep{zhang2024recognize,huang2024tag2text}, object detection \citep{zhu2025revisiting}, concept predictors \citep{oikarinen2023labelfree,konforti2026cb}, or sparse decomposition of image embeddings \citep{choi2025conceptscope,jiang2025interpretable}. We evaluate captioning and tagging approaches in \cref{sec:evaluation} and find that they perform well, but not optimally, and they are costly as they require MLLM inference for a large number of images. Concept-based models with a closed concept set may not include important concepts that are relevant for a dataset, while open concept sets \citep{oikarinen2023labelfree,zhu2022gsclip} depend on the underlying multimodal embedding space, which we investigate in this work.

The second approach (\enquote{propose-then-rank}, PTR) involves sampling small sets of images from the contrasted datasets and asking an MLLM to propose distinguishing features \citep{dunlap2024describing,liu2026organizing}. We show in \cref{sec:evaluation} and \cref{sec:supp:ablations:proposethenrank} that PTR is robust to low prevalence of target concepts, but the same performance or better can be achieved using a generic wordlist. Therefore, we propose to spend the expensive MLLM inference on post-selection optimization, which enables higher gains on complex datasets (\cref{sec:multi-concept}). There, selecting diverse concepts becomes important, because taking the top-$k$ highest-ranking concepts suffers from high redundancy.

\mypara{Dataset Diffing in Text-Only Datasets}
Methods for explaining differences in text-only datasets mainly follow the \enquote{propose-then-rank} approach described above \citep{wang2023goal,dunlap2024describing}. Initially, based on a set of samples, a language model proposes potential explanations, which are sometimes clustered \citep{findeis2024inverse,bravansky2025dataset}. These proposed explanations are then evaluated against the documents to determine which documents are associated with each candidate explanation. Finally, a suitable subset of candidates is selected based on the specific goal, such as clustering \citep{wang2023goal} or \enquote{inverse constitutional AI} \citep{findeis2024inverse}. \citet{zhong2024explaining} propose an additional iterative process to identify better candidate concepts. We show that a similar idea for images works well and enables substantial improvements (\cref{sec:evaluation,sec:multi-concept}).

\mypara{Explaining Distribution Shifts}
Dataset diffing can also be viewed as characterizing distribution shifts, as both tasks assume that datasets differ along a few well-defined attributes. Most work on explaining distribution shifts uses optimal transport (OT). For instance, \citet{kulinski2023towards} use OT to identify a low-dimensional shift direction in multimodal image embedding space. Then, we can interpret the identified direction by retrieving aligned text concepts. \citet{naumann2026wasserstein} propose a similar method but find multiple subspaces. Other work finds a few sample images that best describe the shift, like the most and least shifted samples in OT matchings \citep{brockmeier2021identifying,hulkund2022interpretable} or through image translation \citep{kulinski2022image,kulinski2023towards}. Our experiments in \cref{sec:evaluation} show that subspace-based approaches yield comparatively weak performance, because interpreting subspaces in multimodal embedding spaces is difficult and can introduce noise. Furthermore, OT methods do not trivially generalize to patch-based embedding spaces, because no clear mapping of patches exists across images.

\section{Fine-Grained Dataset Diffing with Multimodal Optimization}
\label{sec:method}

\begin{figure}[t]
    \centering
    \includegraphics[width=\linewidth]{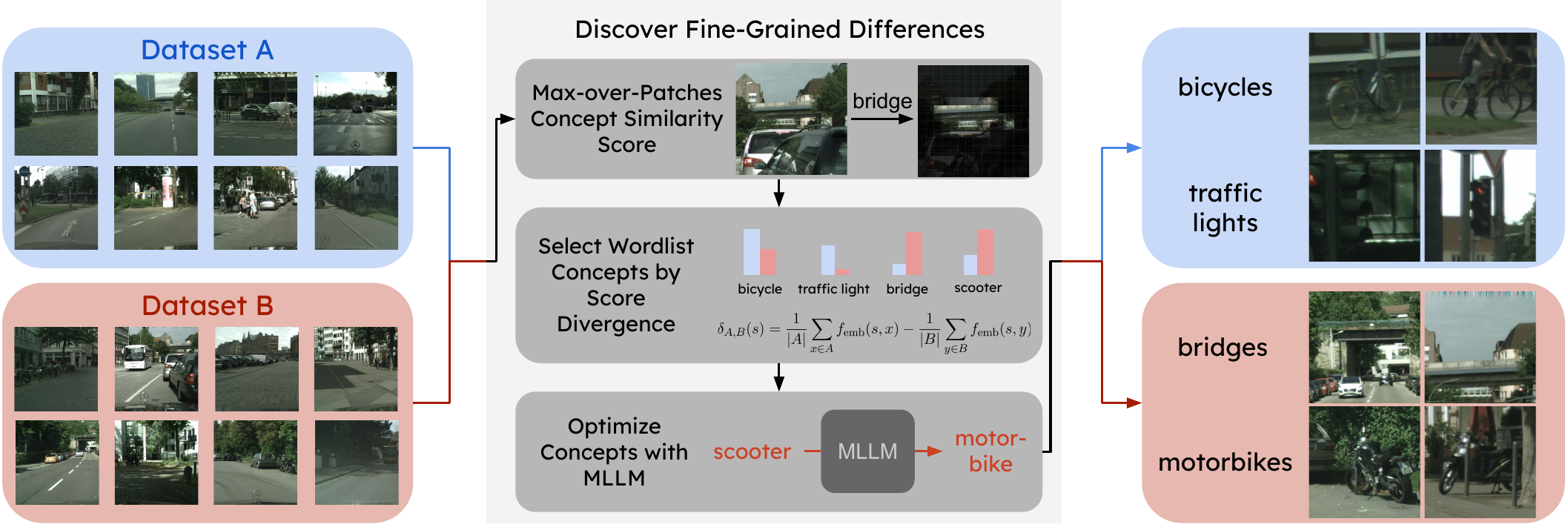}
    \caption{\textbf{How we detect fine-grained differences between datasets.} Left: We are given two sets $A$ and $B$ consisting of natural images. Center: We compute multimodal patch embeddings and calculate max-over-patches cosine similarities to corresponding concept text encodings. Then, we find diverse concepts whose score distribution differs between the sets. Finally, we use an MLLM to refine concepts. Right: The method returns a list of distinguishing concepts for each set.}
\end{figure}

The task we address is defined as follows: Given two sets of images $A$ and $B$, find a list of concepts that together describe the differences between the images in $A$ and the images in $B$, i.e., a list of concepts that are more prevalent in images in $A$ (but not $B$), and vice versa.
This problem can be formalized by the following optimization problem, where we want to find optimal concepts $s^{\ast}$ that maximally distinguish $A$ and $B$ by comparing their prevalence:
\begin{equation}
    s^{\ast} = \arg\max_s \mathcal{R}(s) =  \arg\max_s \mathbb{E}_{x \sim A}[f(s, x)] - \mathbb{E}_{y \sim B}[f(s, y)].
\end{equation}
Here, $s$ is a natural language description of a possible dataset difference and $f$ is a function that scores how relevant $s$ is to an image $x$ or $y$, i.e. higher scores indicate higher relevance. This formulation makes clear how dataset diffing entangles a search problem over language (finding $s$) with a perception problem (evaluating $f$). 

\mypara{Fine-Grained vs.\ Image-Level Embeddings}
One main focus of this paper is the perception function $f$. As $f$ needs to be evaluated many times, for efficiency reasons a vector-similarity-based score is necessary. Concretely, we compare image-level embeddings like SigLIP2 \citep{tschannen2025siglip}, which have been used in prior work, with patch-level multimodal embeddings like FG-CLIP2  \citep{xie2025fg2}. In both cases, let $\phi$ denote the image encoder and $\psi$ the text encoder. We then set
\begin{align}
    f^{\text{CLIP}}_{\mathrm{emb}}(s, x) &= \psi(s)^\top \phi(x) \\
    f^{\text{FG-CLIP}}_{\mathrm{emb}}(s, x) &= \max_{i\in \{1, \ldots, \rho\}} \psi(s)^\top \phi_i(x).
\end{align}
Here, FG-CLIP returns a patch embedding matrix of shape $d \times \rho$, where $\rho$ denotes the number of patches and $\phi_i(x)$ denotes the $i$-th column.

After calculating the $f_{\mathrm{emb}}$ scores for a concept and all images in sets $A$ and $B$, we calculate a divergence score $\delta_{A, B}(s)$ between the score distributions in $A$ and $B$. This score indicates how prevalent the concept is in $A$ in contrast to $B$, and vice versa. Depending on the application, different divergence scores can be chosen (see \cref{sec:supp:methods:fg-clip}). For \cref{sec:evaluation}, we use the difference of means:
\begin{equation}\label{eq:margin}
    \delta_{A,B}(s) = \frac{1}{|A|}\sum_{x \in A} f_{\mathrm{emb}}(s, x)
    - \frac{1}{|B|}\sum_{y \in B} f_{\mathrm{emb}}(s, y).
\end{equation}

Given $M$ candidate concepts $\mathcal{C} = \{s_1, \ldots, s_M\}$, we select $k$ concepts with large $\delta_{A,B}$ as characterizing $A$, and $k$ concepts with low $\delta_{A, B}$ as characterizing $B$. In \cref{sec:evaluation}, we evaluate combining both embedding spaces, and \cref{sec:multi-concept} focuses on concept selection strategies in complex datasets.

\mypara{MLLM-Driven Concept Optimization}
The initial pool of concepts may be suboptimal, and the $k$ selected concepts may be noisy. Therefore, we iteratively optimize the pool using an MLLM; i.e., we implement the search for $s^{\ast}$ as an MLLM-driven optimization problem, similar to previous approaches that utilize LLMs as optimizers \citep{yang2024large,fernando2024promptbreeder}. We provide an MLLM with the $2\times k$ selected concepts and their scores and then request it to generate new candidates that better describe the difference between $A$ and $B$. Specifically, we provide the following information to an MLLM: (i) An image displaying a $2\times 5$ grid of two rows of images selected from sets $A$ and $B$; (ii) For both sets $A$ and $B$, we provide a list of concepts that $f$ scores highly, irrespective of the difference between the two sets. As an initial concept pool to choose from, we use the stopword-filtered COCA corpus wordlist \citep{davies2010coca}, which allows truncating by frequency.

\begin{figure}[t]
\centering
\includegraphics[width=\linewidth]{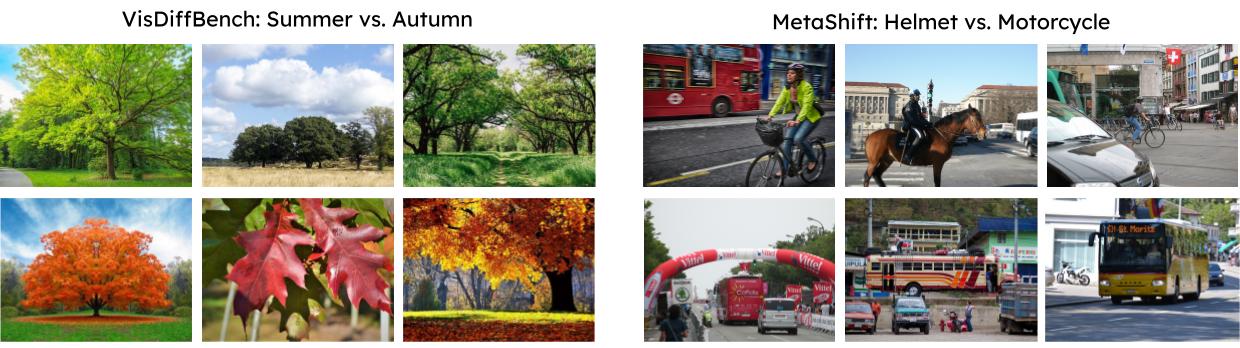}
\caption{\textbf{Example images for VisDiffBench (left) and MetaShift (right).} The summer ($A$, top) and autumn ($B$, bottom) sets differ in a whole-image color shift, whereas the helmet ($A$, top) and motorcycle ($B$, bottom) sets differ in small objects in otherwise similar street scenes.}
\label{fig:visdiffbench_metashift_comparison}
\end{figure}

\section{Evaluation on VisDiffBench and MetaShift}
\label{sec:evaluation}

In the first set of experiments, we get an understanding of which methods perform well. Our main insights are: (1) Standard benchmarks are saturated, as most methods perform well. (2) As expected, patch embeddings are optimal for detecting differences in smaller attributes and objects, and image-level embeddings are optimal for salient themes present in the entire image. (3) Wordlist-based methods already achieve strong performance while being significantly cheaper than MLLM-based concept proposals. MLLM optimization further improves concepts in this setting.

\mypara{Datasets}
We evaluate all methods on VisDiffBench \citep{dunlap2024describing} and MetaShift \citep{liang2022metashift}. VisDiffBench contains 150 subtasks, each consisting of image set pairs ($A$ and $B$), which are organized into three difficulty levels (easy, medium, and hard). Each image set in VisDiffBench contains exactly 100 images.
For MetaShift, we construct a suitable subset of the full benchmark. First, we generate MetaShift for the recommended classes (cat, dog, bus, truck, elephant, horse, bowl, and cup), which results in 212 subtasks. Next, for each image, we use Gemma4-31B to verify the presence of the distinguishing attribute. Images where Gemma4 does not detect the target attribute are removed. Then, we retain the subtasks where both image sets A and B contain at least 20 images each, and we cap the number of images per set at 100. This yields 68 remaining subtasks, with an average of 52.6 images per set. See \cref{fig:visdiffbench_metashift_comparison} for examples from VisDiffBench and MetaShift.

\mypara{Evaluation}
We evaluate descriptions of differences between datasets $A$ and $B$ using three metrics that cover (1) \textit{discrimination} (Disc.), i.e.\ how well the returned concepts separate the datasets. This is measured by the accuracy an MLLM achieves in classifying images into sets $A$ and $B$ based on the concepts. (2) \textit{relevance} (Rel.), i.e.\ whether the concepts actually apply to images in the data. This is measured by the average number of concepts an MLLM attributes to images. (3) \textit{embedding similarity} (Sim.), which measures the maximum cosine embedding similarity between the ground truth concept and any proposed concept. All three metrics are based on $k=10$ concepts returned per side ($A$ and $B$) by each method. The MLLM used for Disc.\ and Rel.\ is Gemma4-31B (\cref{tab:evaluation}), and we include results for Qwen3.8-27B in \cref{sec:supp:ablations:judge} alongside additional details on the metrics.

\mypara{Baselines} 
We evaluate the following baselines: (1) propose-then-rank, (2) distribution shift explanation approaches, and (3) tagging + aggregation methods. For (1) \textit{propose-then-rank}, we follow \citet{dunlap2024describing}, who use an MLLM to propose candidate concepts from $2\times k$ (we use $k=5$) image samples from set $A$ and $B$. We then combine all 10 samples in one composite image and ask an MLLM to propose 10 distinguishing concepts per side, aggregate them over 20 iterations, and rank them with SigLIP or FG-CLIP text-image similarity. See \cref{sec:supp:methods:propose-then-rank} for details. 

For (2) \textit{distribution shift explanations}, we detect subspaces in the SigLIP2 image embedding space and find text concepts that are aligned with the identified subspaces. We evaluate two approaches: U-WaX \citep{naumann2026wasserstein}, which uses optimal transport to find subspaces, and alternatively, we learn orthogonal linear directions from linear support vector classifiers to distinguish between the embeddings of images in both sets. Details are in \cref{sec:supp:methods:distribution-shift}. For (3) \textit{tagging + aggregation methods}, we extract relevant textual information from each image using a captioning model and then use an LLM to aggregate the resulting information into candidate concepts. We evaluate three tagging variations: (a) generating a long-form caption for each image; (b) extracting a structured list of objects and their attributes; and (c) extracting visual concepts using the RAM++ model \citep{zhang2024recognize}. Because RAM++ operates on a fixed label space, here we directly select the 10 concepts with the largest count difference between sets $A$ and $B$. See \cref{sec:supp:methods:tagging} for details.

\mypara{Methods} We evaluate our proposed embedding-based approach (\cref{sec:method}), which scores the stopword-filtered COCA wordlist ($M{=}54{,}000$ concepts) against all images using either SigLIP image-level or FG-CLIP patch-level embeddings and selects the $k{=}10$ concepts per side with the largest difference-of-means margin (\cref{eq:margin}). To refine the initial selection, we optionally apply MLLM-driven concept optimization (\cref{sec:method}), which iteratively proposes and evaluates new candidate phrases given visual examples and the current concept pool. We further evaluate a hybrid configuration that combines both embedding spaces by taking the maximum of their calibrated divergence scores (see \cref{sec:supp:methods:fg-clip} for technical details).

\begin{table}[t]
\centering
\resizebox{\textwidth}{!}{
\begin{tabular}{lcccccccccccc}
\toprule
 & \multicolumn{3}{c}{VDB-Easy} & \multicolumn{3}{c}{VDB-Med.} & \multicolumn{3}{c}{VDB-Hard} & \multicolumn{3}{c}{MetaShift} \\
\cmidrule(rl){2-4}
\cmidrule(rl){5-7}
\cmidrule(rl){8-10}
\cmidrule(rl){11-13}
{} & $D$ & $Rel$ & Sim. & $D$ & $Rel$ & Sim. & $D$ & $Rel$ & Sim. & $D$ & $Rel$ & Sim. \\
\midrule
MLLM + SigLIP (PTR) & {0.981} & {0.809} & \textbf{0.549} & \underline{0.891} & {0.873} & \underline{0.425} & 0.815 & {0.899} & 0.331 & \underline{0.891} & {0.601} & 0.441 \\
MLLM + FG-CLIP (PTR) & \textbf{0.983} & \underline{0.853} & \underline{0.523} & \textbf{0.897} & \underline{0.907} & 0.424 & 0.816 & \underline{0.940} & 0.353 & 0.884 & \underline{0.663} & \underline{0.473} \\
\midrule
Distrib.\ Shift: U-WaX & 0.972 & 0.335 & 0.306 & 0.816 & 0.262 & 0.185 & 0.757 & 0.196 & 0.147 & 0.834 & 0.232 & 0.400 \\
Distrib.\ Shift: Linear & 0.969 & 0.347 & 0.319 & 0.842 & 0.307 & 0.200 & 0.752 & 0.221 & 0.157 & 0.844 & 0.239 & 0.405 \\
\midrule
Caption Summary & 0.974 & 0.613 & 0.338 & 0.858 & 0.734 & 0.267 & 0.805 & 0.654 & 0.236 & 0.825 & 0.423 & 0.252 \\
Image Tagging: MLLM & 0.971 & 0.585 & 0.346 & 0.866 & 0.666 & 0.288 & 0.800 & 0.639 & 0.243 & 0.818 & 0.412 & 0.310 \\
Image Tagging: RAM++ & 0.975 & \textbf{0.907} & 0.335 & 0.833 & \textbf{0.908} & 0.262 & 0.719 & \textbf{0.969} & 0.219 & 0.854 & \textbf{0.767} & 0.424 \\
\midrule
COCA + FG-CLIP & 0.963 & 0.548 & 0.283 & 0.844 & 0.532 & 0.203 & 0.739 & 0.458 & 0.134 & \textbf{0.904} & 0.510 & \textbf{0.501} \\
COCA + SigLIP & 0.976 & 0.679 & 0.336 & 0.846 & 0.672 & 0.275 & 0.792 & 0.680 & 0.275 & 0.738 & 0.402 & 0.306 \\
\quad + MLLM Optim. & \underline{0.982} & 0.777 & {0.511} & 0.875 & 0.757 & \textbf{0.445} & \textbf{0.830} & 0.854 & \textbf{0.402} & 0.757 & 0.408 & 0.307 \\
\quad + FG-CLIP & \underline{0.982} & 0.746 & 0.489 & {0.888} & 0.784 & 0.417 & \underline{0.817} & 0.743 & \underline{0.341} & 0.884 & 0.546 & {0.465} \\
\midrule
True Concept & 0.982 & - & - & 0.889 & - & - & 0.843 & - & - & 0.932 & - & - \\
\bottomrule
\end{tabular}
}
\caption{\textbf{Performance comparison on VisDiffBench and MetaShift.} 
We compare the methods across three metrics: discrimination ($D$), relevance ($Rel$), and embedding similarity (Sim). 
Best scores per benchmark and metric are highlighted in bold, and second-best scores are underlined.}
\label{tab:evaluation}
\vspace{-15pt}
\end{table}

\mypara{Results}
As shown in \cref{tab:evaluation}, image-level embeddings (SigLIP) work well on VDB, where distinguishing concepts are visually salient (VDB-Hard: $D=0.792$ vs.\ oracle $D=0.843$), but fail on localized concepts in MetaShift ($D=0.738$ vs.\ oracle $D=0.932$; see \cref{sec:supp:ablations:mllm} for an ablation).
Conversely, patch-level embeddings (FG-CLIP) work well on MetaShift ($D=0.904$) but less well on abstract concepts in VDB (also see \cref{sec:supp:ablations:fine-grained-perception} for a test of FG-CLIP fine-grained classification). MLLM-based concept optimization improves SigLIP on VDB-Hard ($Rec$ from $0.275$ $\rightarrow$ $0.402$), closing the gap to the ground truth. Our hybrid approach (COCA + MLLM Optim.\ + FG-CLIP) balances both representations and achieves strong performance across both benchmarks.

Tagging methods (e.g., RAM++) achieve reasonable discrimination performance, which confirms the need for localized detection, but MLLM-based captioning per image is prohibitively expensive. Distribution shift baselines (U-WaX, Linear) consistently underperform (Sim.\ $\le 0.200$), which indicates that interpreting geometric shifts in language yields noisy concepts even in multimodal spaces.

Finally, PTR performs strongly overall ($D=0.891$ on MetaShift). On the one hand, its success reveals a fundamental benchmark limitation: shifts are so isolated and consistent that observing the entire distribution is unnecessary, and we only need to look at a few images. On the other hand, in \cref{sec:supp:ablations:proposethenrank} we show that PTR reliably detects even low-prevalence concepts. So the actual question is about cost. Here, we find that wordlist-based methods achieve very similar performance without any MLLM inference, and surpass MLLM proposals when inference is spent on post-selection optimization. However, real-world datasets are more complex than the single-target, high-prevalence shifts modeled in VisDiffBench and MetaShift. Therefore, we design an experiment to explicitly test methods in noisy and complex datasets.

\section{Evaluation on Noisy and Multi-Target Datasets}
\label{sec:multi-concept}

\mypara{Setting and Dataset Construction}
The results in \cref{sec:evaluation} above show that, depending on the dataset, different embedding models are necessary to obtain optimal performance. In particular, in real datasets, generally more than one concept determines the difference, and concepts may not be exclusively tied to one set. Therefore, we construct a harder setting that involves multiple concepts. Given a dataset that has image-level concept annotations, i.e., indicators of which concepts are present in the image, we assign a role for each concept. A concept can be aligned to Set $A$, aligned to Set $B$, or neutral. For aligned concepts, between 75\% and 95\% of images with the concept are in the respective set, and for neutral concepts, at most 60\% of images are in either set. Every class is additionally assigned an overall prevalence that determines how frequent the concept is in the entire dataset ($A$ and $B$ combined). This construction directly addresses the limitations we mentioned earlier: It includes multiple concepts per side, and even for aligned concepts, a relevant portion can still be present in the other set. For technical details on the dataset sampling, see \cref{sec:supp:multi-target:sampling}.

\mypara{Evaluation}
In \cref{sec:evaluation}, we have shown that methods find discriminative and relevant concepts. However, in the complex and noisy datasets that we evaluate in this section, due to the diversity of the target concepts, both are less useful, and even ground-truth concepts don't discriminate sets $A$ and $B$ well in the classification sense (see \cref{sec:supp:multi-concept:metrics} and \cref{tab:supp:multi-concept:discrimination}).

Instead, we focus on recall, i.e., whether methods can recover the ground-truth-aligned concepts. We measure this from three perspectives: (1) \textit{MCC-R} measures how well each ground-truth concept is recovered by any predicted concept. For this, we calculate the maximum Matthews Correlation Coefficient (MCC) across all possible binarization thresholds $t$ on the image-concept similarity scores. (2) \textit{Sim.} measures the maximum cosine embedding similarity between a ground-truth concept and any predicted concept. (3) \textit{$R_J$} uses an LLM to select the ground-truth concepts that are implied by any predicted concept and then calculates recall on this selection.

For MCC, a ground-truth concept is considered recovered if any predicted concept achieves MCC $\geq 0.5$. Intuitively, MCC-R measures how well the predicted structure of the data (predicted concepts + scores) that a user would see matches the ground-truth structure induced by ground-truth concepts and binary presence indicators. See \cref{sec:supp:multi-concept:metrics} for details on the metrics.

\mypara{Methods}
For this experiment, we focus on three contrasts: \textbf{(1)} How do SigLIP and FG-CLIP compare in complex datasets? \textbf{(2)} Can static wordlists compete with MLLM proposals? \textbf{(3)} How can we optimally select and improve the concepts? Accordingly, on all datasets, we compare MLLM-proposed concepts (100 times $2\times 10$ proposals, i.e., 10 concepts for each side for 100 $2\times 5$ image samples) vs.\ the static COCA wordlist, SigLIP vs.\ FG-CLIP $f_{\mathrm{emb}}$ similarity scores used to rank concepts, and we show that suitable concept selection strategies yield significant gains.
Our main insight is that in this setting, simply selecting the top-10 highly ranked concepts is not optimal, because this leads to high redundancy in the selected concepts. We propose the following modifications to the selection process that improve performance: 

\textit{(Multi-Level Masking)} We set all scores for a single image except the top-$K$ to zero. Here, we evaluate three different levels, $K=10$, $K=50$, and $K=100$, and we select the top-10 concepts for each level per side ($A$ and $B$) and reduce them to 10 in a round-robin fashion (see \cref{sec:supp:multi-target:multilevelmasking} for an ablation on the values for $K$). Intuitively, this allows the method to select concepts at different saliency levels. See \cref{sec:supp:multi-target:maskedselection} for details.

\textit{(Redundancy-Aware Selection)} To further eliminate redundancy in naive top-$k$ selection, we select a diverse subset of concepts using Orthogonal Matching Pursuit (Residual-OMP). This greedily chooses concepts that best capture the unexplained variance in the binary set membership labels (in $A$ or in $B$), i.e., at each iteration, Residual-OMP selects the concept that is most correlated with the current residual, and then updates this residual by orthogonalizing the target labels against the subspace spanned by all previously selected concepts. See \cref{sec:supp:multi-target:residual-omp} for details.

\textit{(Optimization)} We use iterative MLLM optimization to find better concepts; see \cref{sec:method}. Furthermore, here, we combine optimization with Residual-OMP. We prioritize MLLM-proposed concepts by adding a constant to their correlation score in Residual-OMP selection. This ensures that MLLM proposals are more likely to be included in the returned concepts.

\begin{table}[t]
\centering
\resizebox{\textwidth}{!}{%
\begin{tabular}{l@{\hspace{8pt}}ccc@{\hspace{8pt}}ccc@{\hspace{8pt}}ccc@{\hspace{8pt}}ccc}
\toprule
& \multicolumn{3}{c}{LVIS} & \multicolumn{3}{c}{MS COCO} & \multicolumn{3}{c}{SUN RGB-D} & \multicolumn{3}{c}{DOTA} \\
\cmidrule(lr){2-4}\cmidrule(lr){5-7}\cmidrule(lr){8-10}\cmidrule(lr){11-13}
{} & MCC-R & Sim. & R$_{\mathrm{J}}$ & MCC-R & Sim. & R$_{\mathrm{J}}$ & MCC-R & Sim. & R$_{\mathrm{J}}$ & MCC-R & Sim. & R$_{\mathrm{J}}$ \\
\midrule
RAM++ & 31.5 & 56.9 & {35.9} & 44.7 & 62.8 & 44.1 & 15.1 & \underline{60.5} & 39.7 & 29.4 & 53.6 & 31.4 \\
\midrule
MLLM + SigLIP (PTR) & 26.4 & 41.9 & 27.1 & 40.2 & 48.5 & 37.3 & 10.1 & 42.6 & 23.5 & 33.8 & 41.7 & 30.2 \\
MLLM + FG-CLIP (PTR) & 22.2 & 39.8 & 23.9 & 27.4 & 40.6 & 25.9 & 15.8 & 42.1 & 27.4 & 31.1 & 40.8 & 27.0 \\
\quad + Multi-Level Masking & 27.5 & 43.1 & 27.5 & 42.4 & 49.4 & 38.2 & 20.6 & 47.7 & 37.3 & 32.4 & 40.0 & 27.6 \\
\quad + Residual-OMP & 32.8 & 45.8 & 33.2 & 53.5 & 54.2 & 48.5 & 22.9 & 49.9 & 44.2 & 45.5 & 47.0 & 41.1 \\
\quad + MLLM Optimization & \underline{39.6} & \underline{61.0} & \underline{42.9} & \textbf{68.0} & \textbf{75.7} & \textbf{66.4} & \underline{24.3} & 58.0 & \underline{46.9} & 46.3 & \underline{55.2} & \underline{43.7} \\
\midrule
COCA + SigLIP & 24.3 & 43.9 & 20.5 & 40.7 & 52.1 & 32.2 & 6.6 & 46.8 & 18.2 & 34.9 & 45.7 & 29.5 \\
COCA + FG-CLIP & 16.5 & 40.6 & 15.1 & 17.8 & 39.5 & 13.1 & 11.3 & 45.5 & 18.8 & 28.2 & 37.9 & 15.9 \\
\quad + Multi-Level Masking & 30.0 & 52.9 & 30.9 & 46.8 & 59.2 & 42.3 & 21.4 & 58.8 & 37.6 & 35.5 & 47.6 & 31.2 \\
\quad + Residual-OMP & 33.4 & 54.6 & 33.1 & 52.7 & 61.1 & 46.1 & 22.6 & 59.3 & 40.0 & \underline{47.7} & 51.1 & 42.3 \\
\quad + MLLM Optimization & \textbf{40.4} & \textbf{62.1} & \textbf{43.1} & \underline{66.7} & \underline{74.3} & \underline{63.1} & \textbf{25.7} & \textbf{62.2} & \textbf{49.7} & \textbf{49.7} & \textbf{63.0} & \textbf{52.1} \\
\midrule
Ground-Truth Concepts & 59.4 & -- & -- & 96.3 & -- & -- & 36.8 & -- & -- & 58.4 & -- & -- \\
Best 10 from COCA & 65.8 & -- & -- & 96.3 & -- & -- & 41.7 & -- & -- & 76.3 & -- & -- \\
\bottomrule
\end{tabular}
}
\caption{\textbf{Evaluation on noisy and multi-target datasets.} We compare concept discovery using different proposal sources (MLLM vs.\ COCA wordlist) and embedding models (SigLIP vs.\ FG-CLIP), and we progressively add our proposed Multi-Level Masking, Residual-OMP, and MLLM Optimization. \textbf{Bold} and \underline{underlined} denote the best and second-best results, respectively.}
\label{tab:complex_data_evaluation_recall}
\end{table}

\mypara{Results}
We evaluate on MS COCO \citep{lin2014microsoft}, LVIS \citep{gupta2019lvis}, SUN RGB-D \citep{song2015sun}, and DOTA \citep{xia2018dota}. For each dataset, we vary the number of images per set (150, 300, 500, and 1000) and the number of aligned classes per set (1 to 7), i.e., all 49 combinations of aligned classes.

Results are in \cref{tab:complex_data_evaluation_recall}. We observe that using fine-grained (FG-CLIP) embeddings requires multi-level masking to be effective. Without masking, FG-CLIP consistently underperforms SigLIP (e.g., $17.8\%$ vs.\ $40.7\%$ MCC-R for COCA-20k on MS COCO); however, when masking is applied, it outperforms SigLIP considerably. Also, we find that our proposed concept selection and optimization yield great improvements on all datasets over simple top-$k$ selection. Residual-OMP improves MCC-R by up to $+11.1\%$ (MS COCO, MLLM).  MLLM optimization further improves MCC-R on MS COCO to $68.0\%$ and $R_{\mathrm{J}}$ to $66.4\%$. Finally, regarding the source of proposals, the static COCA wordlist is competitive against the more expensive MLLM-generated concept pool. Under the optimal configuration (Masking + Residual-OMP + Opt.), COCA outperforms MLLM proposals on LVIS, SUN RGB-D, and DOTA (e.g., $49.7\%$ vs.\ $46.3\%$ MCC-R on DOTA), while MLLM generation maintains a slight advantage only on MS COCO. Overall, these results show that finding distinguishing concepts in our complex and noisy datasets is very challenging, but our proposed methods substantially improve upon baselines. However, there remains a substantial gap to ground-truth concepts, which implies that this problem, unlike benchmarks in \cref{sec:evaluation}, is not yet solved.

\section{Applications: Finding Spurious Correlations in ImageNet and Demographic Biases in T2I Models}
\label{sec:applications}

\mypara{Finding Spurious Correlations in ImageNet}
Prior work has shown that ImageNet contains class-associated visual features and spurious correlations that can be exploited by classifiers
\citep{singla2022salient,neuhaus2023spurious,bykov2023finding}.
We use fine-grained image representations and image dataset diffing to systematically detect such class-concept correlations in ImageNet. Our approach involves multiple stages: First, we select concepts that activate more strongly for a class than for other classes, while filtering concepts that are semantically very similar to the class name. Then, we use an LLM to filter class-concept pairs based on interestingness and plausibility, removing low-quality or obvious correlations, such as ram (class) and pasture (concept). The LLM assigns scores between 1 and 10 for how plausible and interesting a class-concept correlation would be. Finally, we verify the remaining candidates with an MLLM.

To filter concepts that activate more for a class than for other classes, we use dataset diffing in two stages. The first calculates the one-vs-rest AUROC, i.e., how well the concept separates the class from all other classes based on the FG-CLIP image-concept similarity scores. The second filter calculates the AUROC against the 20 classes with the most similar mean $f_{\mathrm{emb}}$ score vectors by cosine similarity. The second filter is necessary to eliminate associations such as dog-collar. \enquote{Collar} is discriminative for many dog classes relative to the entire dataset, but not discriminative for any dog class individually. Overall, for this application, we manually tuned the workflow to yield a manageable set of potentially interesting or unexpected correlations in ImageNet, not all correlations that a classifier could exploit. However, it can be adapted to new datasets or a higher number of candidates. See \cref{sec:supp:applications:imagenet} for more details.

\begin{figure}[t]
\centering
\includegraphics[width=\linewidth]{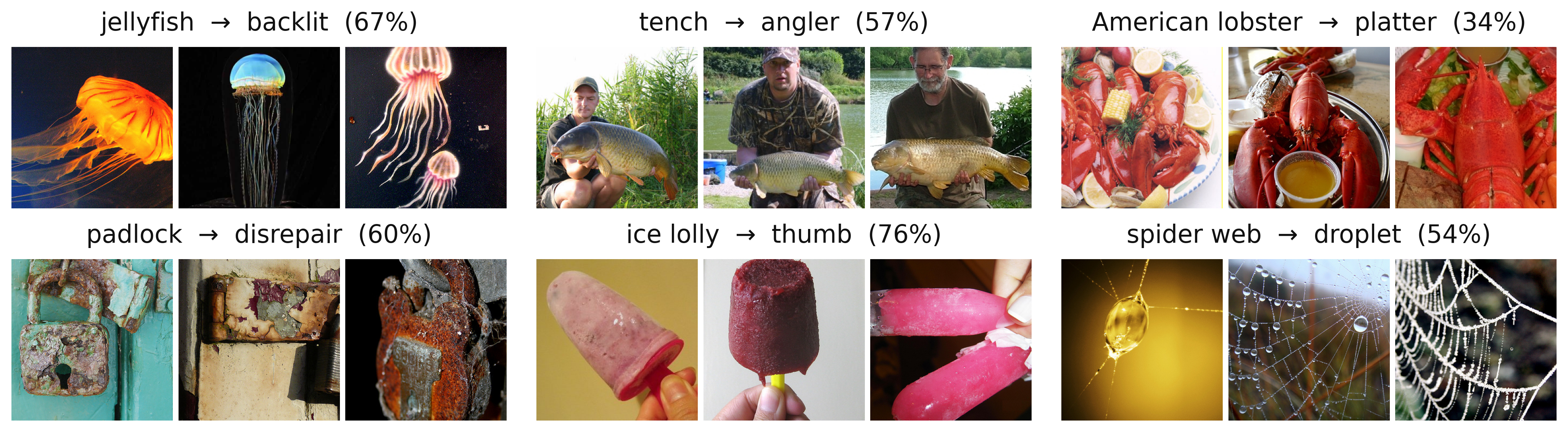}
\caption{\textbf{Examples of spurious class-concept correlations in ImageNet.} For each class $\rightarrow$ concept pair, we show the ratio of class images containing the concept (according to Qwen3.8-27B) with three representative samples.}
\label{fig:imagenet-spurious-correlations}
\end{figure}

Our results are as follows: We tested 6,103 candidate class-concept pairs in MLLM verification. Of these, 1,553 (25.4\%) candidates across 578 classes survived, i.e., the concept is present in at least 10\% of the class' images. We find a number of trends that involve multiple classes: fish presented as catch, i.e., held by a person (e.g., \enquote{tench} $\rightarrow$ \enquote{angler}); animals in zoo/captivity settings (e.g., \enquote{toucan} $\rightarrow$ \enquote{cage}); broken/decaying objects (e.g., \enquote{padlock} $\rightarrow$ \enquote{decrepit}, \enquote{gas pump} $\rightarrow$ \enquote{rundown}); sea animals as seafood (e.g., \enquote{American lobster} $\rightarrow$ \enquote{boiled}); sports equipment with people or body parts (e.g., \enquote{rugby ball} $\rightarrow$ elbow). Furthermore, we find many individually interesting correlations such as sliced figs, graffiti on freight trains, or a surprisingly high number of flags with stars in the \enquote{flagpole} class. We show some examples in \cref{fig:imagenet-spurious-correlations} and more in \cref{sec:supp:applications:additional-results:imagenet}.

Finally, we also run the entire workflow with SigLIP2 $f_{\mathrm{emb}}$ scores instead. SigLIP2 proposed 9,853 candidates, of which the MLLM verified 1,455 (14.8\%, lower than the 25.4\% acceptance rate for FG-CLIP). 609 concepts are proposed and verified both using SigLIP2 and FG-CLIP, but only 43.4\% of FG-CLIP candidates (before verification) are included in SigLIP candidates. As expected, the omitted candidates concentrate on smaller attributes and objects, and surface textures. These results show that FG-CLIP is more precise in detecting correlations, and the represented themes are reasonably complementary between the models.

\mypara{Finding Demographic Biases in Text-to-Image Models}
We use FG-CLIP patch-level embeddings and image dataset diffing to find demographic biases in text-to-image (T2I) generative models, similar to \citep{dinca2024openbias}. For this, we specifically target country, religion, and race categories and analyze which stereotypes text-to-image models associate with these categories, and how strongly they express them. As a basis, we select 10 countries (Egypt, Nigeria, Mexico, Brazil, United States, China, India, Turkey, Germany, Australia), 5 religions (Christianity, Judaism, Islam, Buddhism, Hinduism), and 7 race categories (Native American, Asian, Black, Hispanic, Middle Eastern, Pacific Islander, White). We select 60 occupations and 45 common activities and use two T2I models, Flux 1-dev (Flux) and Stable Diffusion 3.5 Large (SD-3.5), to generate 100 images for every prompt in the format \enquote{\texttt{a\  \{country$\mid$religion$\mid$race adjective\} \{man$\mid$woman\} is \{working as occupation$\mid$doing activity \}}}. The country/religion/race specifier can be empty to create contrast sets.

We then calculate the concept scores $f_{\mathrm{emb}}(s, x)$ using FG-CLIP and SigLIP for the 5,317 visually concrete concepts, and contrast the image sets generated from prompts with country/religion/race specified, and prompts without their specification. We select 30 concepts per contrast set with Multi-Level Masking + Residual-OMP on FG-CLIP $f_{\mathrm{emb}}$ scores, the same method we use in \cref{sec:multi-concept}. Then, we verify concept presence using an MLLM on 1800 ($2\times 900$) images per prompt pair, sampled uniformly across occupation/activity specifiers. See \cref{sec:supp:applications:bias} for further details on how we select and verify concepts.

We notice two general observations: (1) FG-CLIP and SigLIP largely disagree regarding the predicted bias concepts before MLLM verification, as their mean Jaccard index is only 2.4\%. (2) The number of concepts that survive MLLM filtering is moderate for FG-CLIP (25.6\%) and low for SigLIP (10.0\%). This is also due to the strict difference threshold (10\% between sets), which suppresses concepts with different distributions but low overall prevalence. Here, FG-CLIP still predicts more meaningful candidates; i.e., for 85\% of FG-CLIP candidates, the MLLM attribution directions (whether an attribute is more frequent in images with country/religion/race specified or in images from the base distribution) agree with the FG-CLIP direction, but they agree for SigLIP candidates only 53\% of the time. This makes SigLIP bias candidates effectively guesses. 

\begin{figure}[t]
\centering
\includegraphics[width=\linewidth]{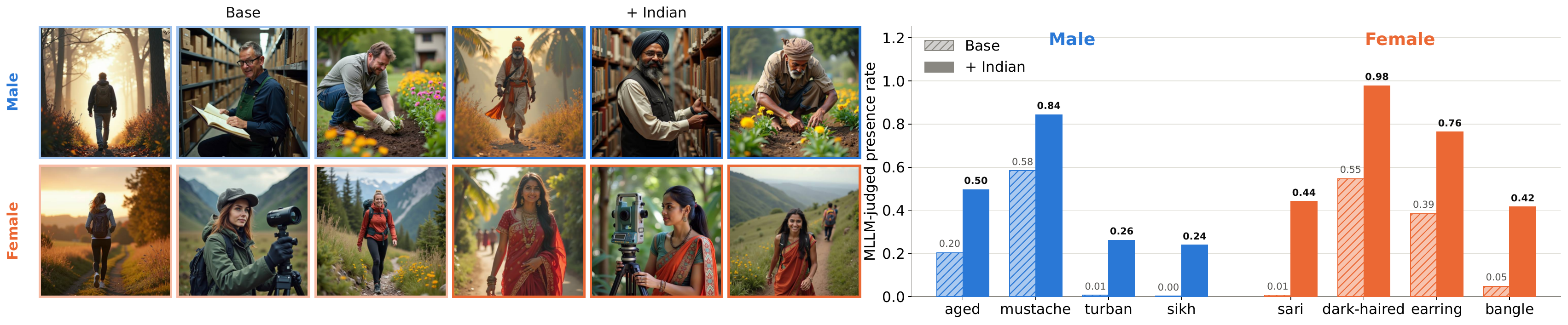}
\caption{\textbf{Demographic bias detection example (\enquote{Indian}).} \emph{Left:} Generated images comparing base prompts against prompts with \enquote{+ Indian} across male (top) and female (bottom) subjects. \emph{Right:} MLLM-judged presence rates for top discovered concepts in base (hatched) versus \enquote{+ Indian}.}
\label{fig:demographic-bias-india}
\end{figure}

We also compare Flux and SD-3.5, where in total we find 407 biased concepts for Flux and 541 for SD-3.5, i.e., SD-3.5 expresses more biases than Flux, while their strengths are comparable: the mean MLLM detection rate difference between prompts is $29.6\%$ for SD-3.5 and $29.5\%$ for Flux. Confirmed concepts largely differ (mean Jaccard index 25.4\%). An example for \enquote{Indian} (country) is in \cref{fig:demographic-bias-india}. There, we find that Indian men are portrayed as older, more bearded, and wearing turbans, while Indian women are portrayed as black-haired and wearing Saris and jewelry. In \cref{sec:supp:applications:demographicbias}, we present more examples.

\section{Conclusion}
\label{sec:conclusion}

Our overarching goal is to understand differences between data distributions, specifically image distributions. In prior work, we found that methods typically rely on MLLM-proposed concepts, which we find costly, and on image-level embeddings, which are known for not representing details well. Intuitively, starting with a broad set of concepts that is cheap to evaluate and first fitting it to the current dataset should be more effective than proposing a potentially large number of concepts from small image samples. Therefore, in this paper, we evaluated and confirmed that MLLM proposals can be replaced by a static wordlist, and in many cases, patch embeddings have clear advantages over image-level embeddings. Furthermore, we found that standard benchmarks are simplistic and do not reflect real-world distributions. Therefore, we designed a setting that is more challenging, i.e., distributions are distinguished by multiple concepts with individually lower prevalence, and discovered that redundancy-aware concept selection, which has not been considered in prior work, is critical in this setting. Finally, we are interested in MLLM-driven optimization for understanding distributional differences and demonstrated that it can improve concept recall performance considerably. By communicating these findings, we significantly advance the field and the understanding of effective methods to interpret distributional differences. Beyond the methodological contributions, we also show how our insights enable useful applications with high practical value, for example, detecting spurious class-concept correlations or comparing outputs of text-to-image models.

\newpage

\subsection*{AI use statement}
In this work, we used generative AI tools for generating synthetic data sets, helping develop theoretical models or conceptual frameworks, proposing or refining hypotheses, designing or providing feedback on research methodology or experiments, implementing methods, cleaning and reformatting datasets, supporting qualitative and thematic data analysis, and interpreting results. We have not used generative AI tools for formulating mathematical claims, assisting with translation, or assisting in the writing of proofs is not applicable to this work. Additionally, we used generative AI tools for suggesting experimental parameters, creating or editing software code, drafting parts of the paper, brainstorming, searching for information, and identifying relevant literature. We have reviewed all AI-assisted work. All cited literature has been read by the authors, and all AI-drafted text and code have been checked and revised by the authors. We take responsibility for the final content of this work, including text, claims, or artifacts produced with the aid of generative AI.

\subsection*{Ethics statement}

This paper is concerned with improving methods that detect distributional differences between image datasets. We do not anticipate any particular concerns that arise from this research. Like all statistical models, embedding models and MLLMs can make mistakes, so outputs should be checked and verified by humans. We ablated all particular model choices for evaluation metrics with alternative models to mitigate artifacts from specific models.

Our experiment on demographic biases involves sensitive attributes such as race or religion. We acknowledge that AI models, including the multimodal embedding models and MLLMs that we use for bias detection, learn skewed representations and associations that may surface in downstream tasks, including bias detection. Therefore, in any deployment and downstream application, outputs must be checked. We also emphasize that attribute labels only concern \textit{perceived} properties, because attributes like gender, race, or religion must be self-identified. Harms from this are mitigated by only labelling synthetic data, not images of real persons. Finally, we acknowledge that some demographic labels, especially race, are social constructs that have no equivalent in nature. We nonetheless include them in our analysis, as even social constructs can cause real-world harm.


\subsection*{Acknowledgements}
This work was partially funded by the ERC (853489 - DEXIM) and the Alfried Krupp von Bohlen und Halbach Foundation, which we thank for their generous support. The authors gratefully acknowledge the scientific support and resources of the AI service infrastructure \textit{LRZ AI Systems} provided by the Leibniz Supercomputing Centre (LRZ) of the Bavarian Academy of Sciences and Humanities (BAdW), funded by Bayerisches Staatsministerium für Wissenschaft und Kunst (StMWK).
The authors gratefully acknowledge the Gauss Centre for Supercomputing e.V. (www.gausscentre.eu) for funding this project by providing computing time on the Supercomputer JUPITER at Jülich Supercomputing Centre (JSC). This work was supported by the EuroHPC JU and the Gauss Centre for Supercomputing (GCS) through funding by the European Commission, German Federal Ministry of Research, Technology and Space (BMFTR), the Ministry of Culture and Science of the State of North Rhine-Westphalia (MKW).
The authors also acknowledge the use of the HPC cluster at Helmholtz Munich for the computational resources used in this study.

\bibliography{iclr2026_conference}
\bibliographystyle{iclr2026_conference}

\newpage
\appendix
\input{supplementary}

\end{document}

%% file: math_commands.tex
\usepackage{amsmath,amsfonts,bm}

\def\eqref#1{equation~\ref{#1}}

\def\1{\bm{1}}

\DeclareMathAlphabet{\mathsfit}{\encodingdefault}{\sfdefault}{m}{sl}
\SetMathAlphabet{\mathsfit}{bold}{\encodingdefault}{\sfdefault}{bx}{n}



%% file: supplementary.tex
\part*{Supplementary Material}

\section{VisDiffbench and MetaShift: Additional Ablations}
\label{sec:supp:ablations}

\subsection{Detailed Description of Metrics and Results with Qwen3.8-27B as Judge}
\label{sec:supp:ablations:judge}

We evaluate methods and baselines in \cref{sec:evaluation} (\cref{tab:evaluation}) using three metrics:

\mypara{Discrimination ($D$)} To evaluate how informative the predicted set of concepts is regarding the difference between the sets $A$ and $B$, we use an MLLM to classify all images ($A$ and $B$ combined) into the sets based on the $2 \times 10$ predicted concepts. For each image, we provide the MLLM with the image, the predicted concepts, and ask it to decide, according to the concepts, whether the image is in set $A$ or in set $B$. After obtaining all per-image predictions, we calculate accuracy as the ratio of correct decisions.

\mypara{Relevance ($Rel.$)} To evaluate whether the predicted concepts actually apply to the images in the data, we use an MLLM to verify the presence of the $2 \times 10$ predicted concepts across the dataset. For each image, we provide the MLLM with the image and the list of predicted concepts, and ask it to output a boolean value for each concept indicating whether it is present in the image. After obtaining all per-image concept assignments, we calculate relevance as the average number of concepts the MLLM attributes to the images.

\mypara{Embedding Similarity (Sim.)} We calculate the maximum cosine embedding similarity between the ground truth concept and all predicted concepts for each side $A$ and $B$, respectively. Then, we average the two per-side similarities. As an embedding model, we use the all-MiniLM-L6-v2 model from \citep{reimers2019sentence}.

In \cref{sec:evaluation} (\cref{tab:evaluation}), we evaluate Discrimination and Relevance using Gemma4-31B as MLLM. Here, we additionally report results for Qwen3.8-27B (in FP8 quantization). They are in \cref{tab:supp:evaluation-qwen} and generally confirm our findings, i.e., they are not contingent on the judge model. Overall, we find that Qwen is slightly stricter in assigning attributes to images, so the relevance metric is shifted down slightly (mean $\Delta=-0.037$, max.\ $\Delta=0.09$). Discrimination metrics stay mostly unchanged (mean $\Delta=0.018$, max.\ $\Delta=0.062$). For VisDiffBench, method rankings are preserved. One interesting finding is that on VDB-Hard, discrimination scores for Propose-then-rank, Embedding Sim.\ + Multimodal Optim., and True Concept are closer to each other, making their difference less notable. On the other hand, Embedding Sim.\ (without optimization) appears comparatively weaker.

For MetaShift, we see changes in ranking, but they appear among the already strong methods, i.e., Propose-then-rank, FG-CLIP, and Attention-Probing. Again, Attention-Probing without optimization appears comparatively weaker than under the Gemma4 evaluation.

\begin{table}[h]
\centering
\resizebox{\textwidth}{!}{
\begin{tabular}{lcccccccccccc}
\toprule
 & \multicolumn{3}{c}{VDB-Easy} & \multicolumn{3}{c}{VDB-Med.} & \multicolumn{3}{c}{VDB-Hard} & \multicolumn{3}{c}{MetaShift} \\
\cmidrule(rl){2-4}
\cmidrule(rl){5-7}
\cmidrule(rl){8-10}
\cmidrule(rl){11-13}
{} & $D$ & $Rel$ & $Rec$ & $D$ & $Rel$ & $Rec$ & $D$ & $Rel$ & $Rec$ & $D$ & $Rel$ & $Rec$ \\
\midrule
MLLM + SigLIP (PTR) & 0.978 & \underline{0.756} & \textbf{0.549} & \textbf{0.880} & \underline{0.807} & \underline{0.425} & \underline{0.806} & \underline{0.845} & 0.331 & 0.863 & \underline{0.561} & 0.441 \\
\midrule
Distrib.\ Shift: U-WaX & 0.963 & 0.313 & 0.306 & 0.799 & 0.258 & 0.185 & 0.714 & 0.214 & 0.147 & 0.805 & 0.229 & 0.400 \\
Distrib.\ Shift: Linear & 0.968 & 0.317 & 0.319 & 0.832 & 0.297 & 0.200 & 0.716 & 0.244 & 0.157 & 0.798 & 0.233 & 0.405 \\
\midrule
Caption Summary & 0.968 & 0.584 & 0.338 & 0.843 & 0.713 & 0.267 & 0.794 & 0.640 & 0.236 & 0.822 & 0.403 & 0.252 \\
Image Tagging: MLLM & 0.968 & 0.552 & 0.346 & 0.858 & 0.635 & 0.288 & 0.783 & 0.608 & 0.243 & 0.817 & 0.392 & 0.310 \\
Image Tagging: RAM++ & 0.972 & \textbf{0.867} & 0.335 & 0.816 & \textbf{0.868} & 0.262 & 0.715 & \textbf{0.931} & 0.219 & 0.838 & \textbf{0.722} & 0.424 \\
\midrule
COCA + FG-CLIP & 0.959 & 0.522 & 0.283 & 0.832 & 0.502 & 0.203 & 0.724 & 0.456 & 0.134 & \textbf{0.878} & 0.472 & \textbf{0.501} \\
COCA + SigLIP & 0.947 & 0.632 & 0.336 & 0.809 & 0.639 & 0.275 & 0.750 & 0.645 & 0.275 & 0.723 & 0.403 & 0.306 \\
\quad + MLLM Optim. & \underline{0.981} & 0.690 & \underline{0.511} & 0.868 & 0.681 & \textbf{0.445} & \textbf{0.808} & 0.769 & \textbf{0.402} & 0.757 & 0.397 & 0.307 \\
\quad + FG-CLIP & \textbf{0.982} & 0.668 & 0.489 & \underline{0.877} & 0.704 & 0.417 & 0.805 & 0.681 & 0.341 & 0.857 & 0.490 & \underline{0.465} \\
\midrule
True Concept & 0.980 & - & - & 0.880 & - & - & 0.818 & - & - & 0.891 & - & - \\
\bottomrule
\end{tabular}
}
\caption{Replication of \cref{tab:evaluation} using Qwen3.8-27B as MLLM judge for $D$ and $Rel$ metrics.}
\label{tab:supp:evaluation-qwen}
\end{table}

\subsection{Ablating the MLLM for SigLIP Embedding Similarity with Optimization}
\label{sec:supp:ablations:mllm}

In \cref{tab:evaluation}, we used Gemma4-31B as MLLM and disabled reasoning for multimodal optimization. While this performs well on VisDiffBench, results on MetaShift are worse than other methods. We attribute this to using image-level SigLIP embeddings, which do not represent the fine-grained details in MetaShift sufficiently well. However, Propose-then-Rank also uses SigLIP embeddings, but performs well also on MetaShift. Therefore, the reason is not only the embedding space, but rather how it interacts with the MLLM. In order to show that this is not an artifact of using Gemma4-31B, we ablate the following variations on MetaShift (Embedding Sim.\ + Multimodal Optim.): (1) Gemma4-31B with reasoning enabled; (2) Gemma4-31B with high-resolution image input; (3) Qwen3.6-27B with reasoning enabled; (4) MiniMax M3 and MiMo-V2.5 as frontier LLMs.

\begin{figure}[ht]
    \centering
    \includegraphics[width=\linewidth]{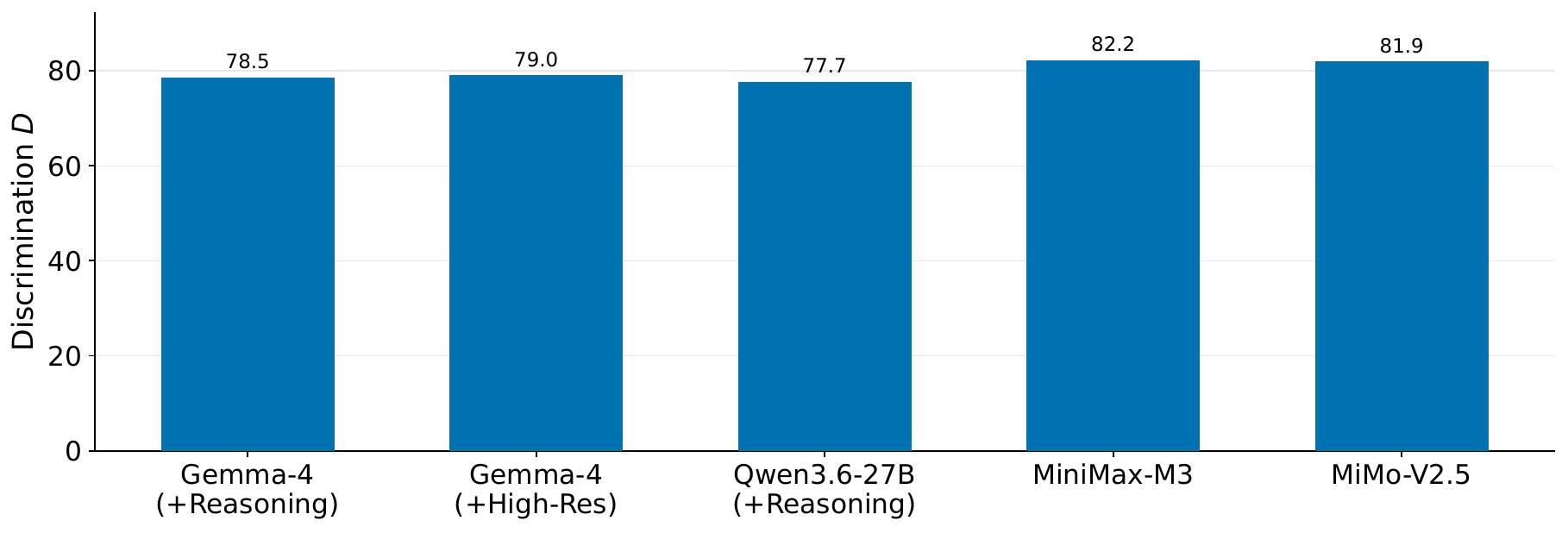}
    \caption{MLLM ablation on MetaShift. Discrimination performance ($D$) of the Embedding Sim.\ + Multimodal Optim.\ method across different MLLMs and inference configurations.}
    \label{fig:supp:metashift-ablation}
\end{figure}

Results for discrimination ($D$, using Gemma4-31B as judge) are in \cref{fig:supp:metashift-ablation}. We see that every setting improves over the baseline value reported in \cref{tab:evaluation} (0.757). However, performance is still worse generally than any other method, even for frontier MLLMs. This shows that the feature initialization with SigLIP embeddings is problematic in a way that any MLLM struggles to recover useful information about the dataset difference.

\subsection{Evaluating Embeddings Models on Fine-Grained Perception}
\label{sec:supp:ablations:fine-grained-perception}

\begin{wrapfigure}{r}{0.5\textwidth}
\centering
  \includegraphics[width=0.48\textwidth]{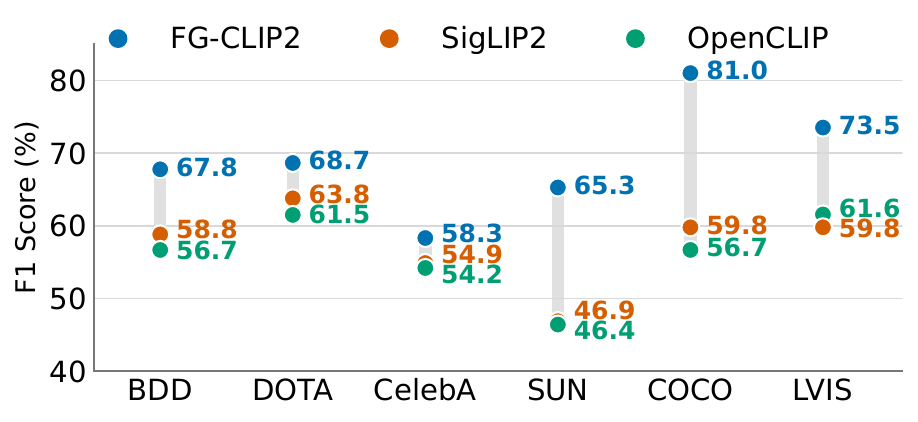}
  \caption{Fine-grained attribute detection performance (F1 score) across six datasets.}
  \label{fig:detectionablation}
\end{wrapfigure}
We analyse the concrete advantage of FG-CLIP in detecting fine-grained attributes. Given a dataset with instance-level annotations, we convert the annotations into image-level presence labels, i.e., whether an attribute or object is present in the image, so the task is a per-attribute binary classification task. We use six datasets that provide such  presence labels derived from their own annotations: BDD100K \citep{yu2020bdd100k} (driving scenes; small road objects and lane-marking types), DOTA
\citep{xia2018dota} (aerial imagery; very small objects in large tiles), CelebA \citep{liu2015deep} (facial attributes, i.e., non-object properties), SUN RGB-D \citep{song2015sun} (cluttered indoor scenes), MS COCO \citep{lin2014microsoft} and LVIS \citep{gupta2019lvis} (common objects). From each dataset, we draw at most $10{,}000$ images and only classes whose prevalence in the sample lies in $[10\%, 90\%]$. The $10,000$ images are selected in a way that maximizes the number of classes that fulfill this criterion.

For each attribute class, we calculate the F1 score based on class name (as defined in the dataset), and based on the best concept among the 54K words used in \cref{sec:evaluation} according to F1 score. For each model and dataset, we individually calculate the per-attribute threshold on the image-text embedding similarities that maximises the F1 score. We compare three model families, i.e., FG-CLIP, SigLIP2, and OpenCLIP.

Results are in \cref{fig:detectionablation}. On class-name-based classification, the best FG-CLIP2 checkpoint reaches an average F1 of $69.1$, against $57.3$ for
the best SigLIP2 and $56.2$ for the best OpenCLIP model. This demonstrates a clear advantage of FG-CLIP on fine-grained detection. The advantage is most pronounced on object detection datasets like COCO, LVIS, and SUN-RGBD, and less pronounced on CelebA, DOTA, and BDD100K, which include less complex scenes. This means that SigLIP and CLIP models have the ability to represent small objects, but they get increasingly suppressed when scenes get more complex. We conclude that there is a genuine representational difference between FG-CLIP patch embeddings and image-level embeddings, and this difference surfaces in realistic datasets, so we can expect it to appear in reality.

\subsection{Ablating the Prevalence of Target Concepts for Propose-then-Rank}
\label{sec:supp:ablations:proposethenrank}

\cref{tab:evaluation} shows that propose-then-rank (using SigLIP for ranking) performs well on both MetaShift and VisDiffBench. However, in theory, this method's performance scales negatively with the prevalence of target concepts in the data, because it does not use a global concept list for guidance, but derives all concepts from small samples (e.g., $2\times 5$ images) of the data via MLLM proposals. In this case, the probability of seeing an image with the target concept is $1-(1-p)^k$, where $p$ is the prevalence of the target concept, i.e., the number of images in the set that have the concept, and $k$ is the number of images per sample. Additionally, even when the target concept is in a sample, there is no guarantee the MLLM will propose it as a candidate.
Therefore, we analyze propose-then-rank for different prevalence strengths on the ImageNet and LVIS datasets. For all prevalence strengths from 2\% to 50\% in 2\% steps, we sample 5 class pairs from each dataset and construct, for each pair, datasets with 50, 100, 200, 300 and 500 images per set ($A$ and $B$). The sampled classes are the target concepts and become the majority class of their respective set, i.e., no other class is represented more frequently in that set. The remaining images are filled randomly from other classes, with the majority constraint. From each dataset, we sample 100 subsamples of $2\times 5$ images (5 from each set) and let the MLLM propose 10 candidate concepts for each set. These are subsequently ranked with SigLIP2 and FG-CLIP, and we select the 10 highest scoring concepts. We then ask an LLM judge whether a set of candidates (both the per-sample and the final top-ranked candidates) contains the ground-truth class or a closely related expression. From this, we calculate two metrics: The expected hit rate, i.e., the likelihood of predicting the ground-truth class from an individual sample, and the final ranking quality, i.e., whether the ground-truth concept is in the 10 final returned concepts for a set.

\begin{wrapfigure}{r}{0.62\textwidth}
\centering
  \includegraphics[width=0.6\textwidth]{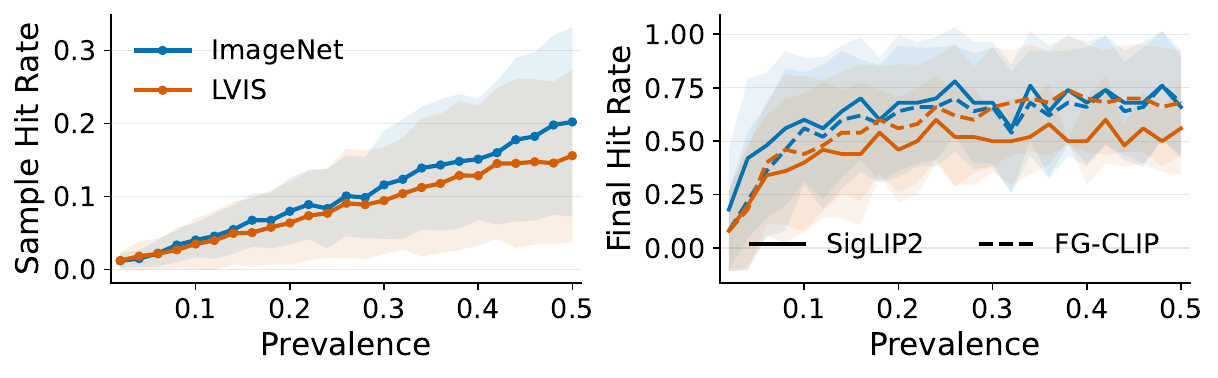}
  \caption{Propose-then-rank effectiveness across target concept prevalences. Shaded regions indicate variance.}
  \label{fig:proposethenrankablation}
\end{wrapfigure}

Results are in \cref{fig:proposethenrankablation}. We find that indeed the hit rate is small for $p < 5\%$, but monotonically increases with prevalence. The final score (after 100 samples) increases sharply below $p < 10\%$ and only slowly increases or stagnates for higher prevalences. This explains the strong performance of propose-then-rank, as for even somewhat prevalent concepts ($\approx$ 10\% of images), the MLLM will find it with enough samples. However, the variance between concepts is high, as some are easier to detect than others, and our judge is rather strict on concept equivalence. Finally, we note that on LVIS, as expected, FG-CLIP yields better concept recall, while SigLIP and FG-CLIP perform similarly on ImageNet with a small advantage for SigLIP. This shows that propose-then-rank can also benefit from fine-grained perception.

\subsection{Testing MLLMs on Fine-Grained Concept Detection}
\label{sec:supp:ablations:mllm-detection}

Throughout this paper, we use MLLMs (Gemma4-31B and Qwen3.8-27B) to evaluate methods, mostly through verifying that concepts are present or absent in given images. To show that Gemma4-31B and Qwen3.8-27B yield reliable concept indicators, we report their classification performance on MS COCO, SUN RGB-D, CelebA, DOTA, and BDD100K. From each dataset, we select classes that appear in at least 10\% of images and in at most 90\% of images. For each class, we sample 500 positive images, i.e., images where the class is present, and 500 negative images, i.e., images where the class is absent, and use the MLLM to predict the class presence.

In \cref{tab:supp:mllm-detection}, we report accuracy, precision, recall, and F1 score. We find that Gemma4-31B and Qwen3.8-27B perform almost identically on all datasets, reaching around 80\% accuracy on average. On BDD100K, DOTA, CelebA, and SUN RGB-D, accuracy is around 80\%. On MS COCO, it is higher, $\approx 90\%$ for Gemma4-31B and $\approx 88\%$ for Qwen3.8-27B. The similar performance across all datasets, including the better performance on MS COCO, indicates that remaining errors are dataset ambiguities or annotation problems instead of systematic MLLM errors.

\begin{table}[h]
\centering
\begin{tabular}{lcccccccc}
\toprule
 & \multicolumn{4}{c}{Gemma4-31B} & \multicolumn{4}{c}{Qwen3.8-27B} \\
\cmidrule(rl){2-5}
\cmidrule(rl){6-9}
Dataset & Acc & Prec & Rec & F1 & Acc & Prec & Rec & F1 \\
\midrule
BDD100K & 0.784 & 0.893 & 0.676 & 0.758 & 0.799 & 0.930 & 0.649 & 0.760 \\
DOTA & 0.778 & 0.930 & 0.619 & 0.721 & 0.788 & 0.948 & 0.610 & 0.730 \\
CelebA & 0.792 & 0.821 & 0.778 & 0.762 & 0.777 & 0.815 & 0.735 & 0.736 \\
SUN RGB-D & 0.810 & 0.818 & 0.828 & 0.813 & 0.808 & 0.839 & 0.785 & 0.802 \\
MS COCO & 0.904 & 0.968 & 0.836 & 0.890 & 0.879 & 0.988 & 0.768 & 0.855 \\
\midrule
Average & 0.800 & 0.863 & 0.747 & 0.775 & 0.797 & 0.877 & 0.711 & 0.763 \\
\bottomrule
\end{tabular}
\caption{MLLM binary detection accuracy, macro-averaged over classes inside each dataset. Evaluation uses up to 500 positive + 500 negative sampled images per class.}
\label{tab:supp:mllm-detection}
\end{table}

\section{Multi-Target Datasets: Additional Ablations}
\label{sec:supp:multi-concept}

\subsection{Ablating Residual-OMP for Redundancy-Aware Concept Selection}
\label{sec:supp:multi-concept:selectionmethod}

In \cref{sec:multi-concept}, we propose to use Residual-OMP (see \cref{sec:supp:multi-target:residual-omp}) to select diverse concepts instead of simply selecting the top-$K$ highly scoring concepts. However, it is not difficult to come up with many variations that fulfill similar purposes. Therefore, in \cref{tab:supp:multi-concept:selectionmethod}, we compare top-$k$ selection and Residual-OMP to a number of similar methods (which we briefly describe below), and find that, while many methods perform similarly well, Residual-OMP is the best method on average across datasets according to both settings (MLLM and COCA wordlist) and all metrics.

\begin{table}[h]
\centering
\resizebox{\textwidth}{!}{%
\begin{tabular}{cl@{\hspace{8pt}}ccc@{\hspace{8pt}}ccc@{\hspace{8pt}}ccc@{\hspace{8pt}}ccc@{\hspace{8pt}}ccc}
\toprule
& & \multicolumn{3}{c}{LVIS} & \multicolumn{3}{c}{MS COCO} & \multicolumn{3}{c}{SUN RGB-D} & \multicolumn{3}{c}{DOTA} & \multicolumn{3}{c}{Avg.} \\
\cmidrule(lr){3-5}\cmidrule(lr){6-8}\cmidrule(lr){9-11}\cmidrule(lr){12-14}\cmidrule(lr){15-17}
& {} & MCC-R & Sim. & R$_{\mathrm{J}}$ & MCC-R & Sim. & R$_{\mathrm{J}}$ & MCC-R & Sim. & R$_{\mathrm{J}}$ & MCC-R & Sim. & R$_{\mathrm{J}}$ & MCC-R & Sim. & R$_{\mathrm{J}}$ \\
\midrule
\multirow{9}{*}{\rotatebox[origin=c]{90}{MLLM}} & Top-$K$ & 27.5 & 43.1 & 27.5 & 42.4 & 49.4 & 38.2 & 20.6 & 47.7 & 37.3 & 32.4 & 40.0 & 27.6 & 30.7 & 45.1 & 32.7 \\
 & MMR & \underline{32.7} & \underline{45.7} & \underline{32.9} & \underline{52.9} & 54.0 & \underline{48.2} & 22.5 & \underline{49.5} & \underline{43.6} & 42.5 & 45.7 & 37.6 & \underline{37.7} & \underline{48.7} & \underline{40.6} \\
 & mRMR & 32.3 & 45.6 & \underline{32.9} & 51.9 & 53.9 & 47.3 & \underline{22.7} & \underline{49.5} & 42.9 & 40.2 & 44.7 & 34.9 & 36.8 & 48.4 & 39.5 \\
 & mRMR cross-level & 30.8 & 44.9 & 32.3 & 50.4 & 52.4 & 45.5 & 21.2 & 49.0 & 42.3 & 39.5 & 44.7 & 34.7 & 35.5 & 47.8 & 38.7 \\
 & Forward Selection & 29.2 & 44.0 & 29.3 & 46.1 & 50.2 & 40.7 & 21.2 & 49.1 & 41.9 & 38.3 & 42.7 & 32.6 & 33.7 & 46.5 & 36.1 \\
 & Cluster-then-Select & 31.0 & 45.0 & 31.7 & 51.6 & 53.3 & 46.2 & 21.4 & 49.4 & 43.4 & \underline{44.1} & \underline{46.0} & 39.2 & 37.1 & 48.4 & 40.1 \\
 & Facility location & 24.3 & 39.2 & 17.5 & 35.7 & 42.4 & 24.7 & 20.3 & 47.4 & 34.5 & 42.8 & 44.9 & \underline{39.4} & 30.8 & 43.5 & 29.0 \\
 & $\ell_1$ LogReg & 31.7 & 45.4 & 32.5 & 52.3 & \underline{54.1} & 48.1 & 22.1 & 49.3 & 42.9 & 42.8 & 45.7 & 37.9 & 37.2 & 48.6 & 40.4 \\
 \cmidrule(l){2-17}
 & {Residual-OMP} & \textbf{32.8} & \textbf{45.8} & \textbf{33.2} & \textbf{53.5} & \textbf{54.2} & \textbf{48.5} & \textbf{22.9} & \textbf{49.9} & \textbf{44.2} & \textbf{45.5} & \textbf{47.0} & \textbf{41.1} & \textbf{38.6} & \textbf{49.2} & \textbf{41.8} \\
\midrule
\multirow{9}{*}{\rotatebox[origin=c]{90}{COCA-20k}} & Top-$K$ & 30.0 & 52.9 & 30.9 & 46.8 & 59.2 & 42.3 & 21.4 & 58.8 & 37.6 & 35.5 & 47.6 & 31.2 & 33.4 & 54.6 & 35.5 \\
 & MMR & 33.1 & 54.2 & 32.4 & 52.3 & 60.9 & 45.9 & 22.0 & 58.9 & 39.5 & 42.9 & 49.9 & 38.3 & 37.6 & 56.0 & 39.0 \\
 & mRMR & \underline{33.3} & \textbf{54.8} & \textbf{33.5} & 52.1 & \underline{61.1} & \underline{46.3} & \underline{22.5} & 59.3 & \underline{40.0} & 41.5 & 49.9 & 36.6 & 37.4 & 56.3 & 39.1 \\
 & mRMR cross-level & 31.0 & 52.8 & 30.9 & 48.5 & 58.3 & 41.6 & 21.1 & 58.2 & 38.0 & 39.7 & 48.6 & 35.2 & 35.1 & 54.5 & 36.4 \\
 & Forward Selection & 31.4 & 53.7 & 31.1 & 46.4 & 57.6 & 38.8 & 21.9 & \textbf{60.3} & \textbf{40.6} & 39.8 & 48.6 & 35.5 & 34.9 & 55.1 & 36.5 \\
 & Cluster-then-Select & 31.7 & 53.3 & 31.1 & 50.0 & 59.6 & 43.8 & 21.1 & 58.3 & 37.4 & 43.5 & 48.6 & 37.2 & 36.6 & 55.0 & 37.4 \\
 & Facility Location & 26.6 & 48.2 & 21.4 & 38.8 & 50.9 & 28.0 & 17.0 & 53.2 & 27.2 & 40.3 & 50.0 & 33.8 & 30.7 & 50.6 & 27.6 \\
 & $\ell_1$ LogReg & 33.0 & \underline{54.6} & \underline{33.3} & \underline{52.5} & \textbf{61.5} & \textbf{46.8} & \textbf{22.6} & \underline{59.4} & 39.8 & \underline{44.7} & \underline{50.7} & \underline{40.6} & \underline{38.2} & \textbf{56.6} & \underline{40.1} \\
 \cmidrule(l){2-17}
 & {Residual-OMP} & \textbf{33.4} & \underline{54.6} & 33.1 & \textbf{52.7} & \underline{61.1} & 46.1 & \textbf{22.6} & 59.3 & \underline{40.0} & \textbf{47.7} & \textbf{51.1} & \textbf{42.3} & \textbf{39.1} & \underline{56.5} & \textbf{40.4} \\
\bottomrule
\end{tabular}
}
\caption{Comparing Residual-OMP as redundancy-aware selection method to alternative methods.}
\label{tab:supp:multi-concept:selectionmethod}
\end{table}

The compared methods are defined as follows:

\mypara{MMR} Maximal Marginal Relevance iteratively selects concepts by maximizing a trade-off between the candidate's score divergence $\delta_{A,B}(s)$ and its redundancy with already selected concepts. Redundancy is measured as the maximum correlation between the standardized scores of a candidate and those of any previously selected concept.

\mypara{mRMR} Minimum Redundancy Maximum Relevance operates identically to MMR, but computes the redundancy penalty differently. Instead of penalizing the maximum correlation, mRMR penalizes the mean correlation between the candidate and all previously selected concepts.

\mypara{mRMR cross-level} This variant of mRMR is specifically designed to improve the multi-level masked concept selection described in \cref{sec:supp:multi-target:maskedselection}. Instead of selecting concepts independently for each masking level $K$, it carries over the selection history from earlier levels. Candidates chosen at earlier levels can never be re-picked, and the method explicitly penalizes similarity to them, which frees up selection budget for new concepts.

\mypara{Forward Selection} Forward sequential selection iteratively selects the set of concepts to maximize joint predictive performance. At each step, it adds the remaining candidate that yields the highest cross-validated ROC-AUC when combined with the already selected concepts using a logistic regression classifier to predict set membership ($A$ vs.\ $B$).

\mypara{Cluster-then-Select} This method groups the candidate concepts into exactly $k$ clusters based on their pairwise score distances (computed as $1$ minus the correlation) using agglomerative clustering with average linkage. For each cluster, the concept with the highest score divergence $\delta_{A,B}(s)$ is chosen as the representative. This naturally yields $k$ distinct concepts while retaining the most discriminative features within each cluster.

\mypara{Facility Location} This approach models concept selection as a submodular maximization problem based on the facility location objective. It iteratively selects candidates that maximize the marginal gain in covering the similarity matrix of the candidate pool, weighted by the non-negative score divergence of the candidates.

\mypara{$\ell_1$ LogReg} This method fits an $\ell_1$-regularized logistic regression model to predict the binary set membership labels using the standardized score matrix as input. We search over a grid of regularization strengths to find the value that yields exactly $k$ non-zero coefficients. The corresponding concepts are then selected and ordered by the absolute magnitude of their learned weights.

\subsection{Ablating the Number of Predicted Concepts per Side}
\label{sec:supp:multi-concept:conceptnumber}

In \cref{sec:multi-concept}, we evaluate all methods based on 10 predicted concepts per side ($A$ and $B$), i.e., 20 concepts in total. However, this is an arbitrary choice whose main utility lies in its compactness, which makes it easily understandable by human users, and 10 concepts per side exceeds the maximum number of per-side aligned concepts in the dataset construction, which is seven. To validate that our findings are not an artifact of this particular choice of the number of predicted concepts, we evaluate all methods using $k=5,10,15,20,30$ concepts per side. Results are in \cref{tab:supp:multi-concept:conceptnumber} (averaged across datasets). Naturally, recall numbers correlate with the number of selected concepts, but the relative ordering of methods and their margins stay the same.

 \begin{table}[h]
  \centering
  \resizebox{\textwidth}{!}{%
  \begin{tabular}{l@{\hspace{8pt}}ccc@{\hspace{8pt}}ccc@{\hspace{8pt}}ccc@{\hspace{8pt}}ccc@{\hspace{8pt}}ccc}
  \toprule
   & \multicolumn{3}{c}{$k = 5$} & \multicolumn{3}{c}{$k = 10$} & \multicolumn{3}{c}{$k = 15$} & \multicolumn{3}{c}{$k = 20$} & \multicolumn{3}{c}{$k = 30$} \\
  \cmidrule(lr){2-4}\cmidrule(lr){5-7}\cmidrule(lr){8-10}\cmidrule(lr){11-13}\cmidrule(lr){14-16}
  {} & R@$k$ & Sim. & R$_{\mathrm{J}}$ & R@$k$ & Sim. & R$_{\mathrm{J}}$ & R@$k$ & Sim. & R$_{\mathrm{J}}$ & R@$k$ & Sim. & R$_{\mathrm{J}}$ & R@$k$ & Sim. & R$_{\mathrm{J}}$ \\
  \midrule
  MLLM + FG-CLIP & 20.3 & 36.1 & 20.9 & 24.1 & 40.8 & 26.0 & 26.7 & 43.7 & 30.6 & 28.7 & 45.9 & 34.0 & 31.7 & 48.9 & 39.3 \\
  \quad + Multi-Level Masking & 25.6 & 39.7 & 25.4 & 30.7 & 45.1 & 32.7 & 34.1 & 48.4 & 38.5 & 36.6 & 50.8 & 42.7 & 40.6 & 54.3 & 49.8 \\
  \quad + Residual-OMP & 30.5 & 42.1 & 30.0 & 38.6 & 49.2 & 41.8 & 42.4 & 52.8 & 49.2 & 44.6 & 54.9 & 53.6 & 47.6 & 57.9 & 60.3 \\
  \quad + MLLM Optimization & \underline{36.1} & \underline{54.4} & \underline{36.8} & \underline{44.5} & \underline{62.3} & \underline{49.5} & \underline{48.3} & \underline{66.1} & \underline{56.5} &
  \underline{50.7} & \underline{68.5} & \underline{61.4} & \underline{53.3} & \underline{71.6} & \underline{68.0} \\
  \midrule
  COCA + SigLIP & 22.0 & 42.4 & 19.4 & 26.6 & 47.1 & 25.1 & 30.0 & 49.8 & 29.7 & 32.6 & 51.8 & 33.3 & 35.9 & 54.7 & 38.9 \\
  COCA + FG-CLIP & 14.9 & 36.7 & 11.2 & 18.4 & 40.9 & 15.7 & 20.7 & 43.3 & 18.8 & 22.4 & 45.0 & 20.9 & 24.7 & 47.5 & 24.7 \\
  \quad + Multi-Level Masking & 27.4 & 48.7 & 26.9 & 33.4 & 54.6 & 35.5 & 36.9 & 58.0 & 41.2 & 39.6 & 60.5 & 45.8 & 43.6 & 63.9 & 52.5 \\
  \quad + Residual-OMP & 31.4 & 50.2 & 29.7 & 39.1 & 56.5 & 40.4 & 42.9 & 59.5 & 46.4 & 45.3 & 61.5 & 50.6 & 48.4 & 64.0 & 56.7 \\
  \quad + MLLM Optimization & \textbf{36.5} & \textbf{56.9} & \textbf{38.2} & \textbf{45.1} & \textbf{65.0} & \textbf{51.2} & \textbf{49.0} & \textbf{68.9} & \textbf{58.4} & \textbf{51.6} & \textbf{71.5} &
  \textbf{63.5} & \textbf{54.9} & \textbf{74.7} & \textbf{69.7} \\
  \midrule
  Oracle (ground-truth concepts) & 62.7 & -- & -- & 62.7 & -- & -- & 62.7 & -- & -- & 62.7 & -- & -- & 62.7 & -- & -- \\
  Oracle (best $k$ from COCA) & 70.0 & -- & -- & 70.0 & -- & -- & 70.0 & -- & -- & 70.0 & -- & -- & 70.0 & -- & -- \\
  \bottomrule
  \end{tabular}
  }
  \caption{Evaluating methods from \cref{sec:multi-concept} with varying number of predicted concepts per side. Metrics averaged across datasets.}
  \label{tab:supp:multi-concept:conceptnumber}
  \end{table}

\subsection{Ablating the Image Sampling in MLLM Optimization}
\label{sec:supp:multi-target:mllmimagesampling}

In \cref{sec:multi-concept}, we optimize concepts using an MLLM by randomly sampling images from sets $A$ and $B$ and asking the MLLM to propose new concepts that describe the difference between the sets. An alternative is to specifically select images that are not explained by the already selected concepts. Here, we ablate how useful this modification is against sampling random images for the MLLM. The results are in \cref{tab:supp:multi-concept:mllmimagesampling}. We find that both sampling strategies perform similarly, but random sampling performs marginally better. We conclude that optimizing the sampling process for MLLM optimization is nontrivial.

\begin{table}[h]
\centering
\resizebox{\textwidth}{!}{%
\begin{tabular}{l@{\hspace{8pt}}ccc@{\hspace{8pt}}ccc@{\hspace{8pt}}ccc@{\hspace{8pt}}ccc}
\toprule
& \multicolumn{3}{c}{LVIS} & \multicolumn{3}{c}{MS COCO} & \multicolumn{3}{c}{SUN RGB-D} & \multicolumn{3}{c}{DOTA} \\
\cmidrule(lr){2-4}\cmidrule(lr){5-7}\cmidrule(lr){8-10}\cmidrule(lr){11-13}
{} & MCC-R & Sim. & R$_{\mathrm{J}}$ & MCC-R & Sim. & R$_{\mathrm{J}}$ & MCC-R & Sim. & R$_{\mathrm{J}}$ & MCC-R & Sim. & R$_{\mathrm{J}}$ \\
\midrule
MLLM + Optimization & 39.6 & 61.0 & 42.9 & 68.0 & 75.7 & 66.4 & 24.3 & 58.0 & 46.9 & 46.3 & 55.2 & 43.7 \\
\quad + unexplained images & {39.4} & {60.4} & {42.5} & {67.6} & {75.2} & {65.3} & {24.2} & 57.8 & {47.1} & 46.2 & {54.9} & {43.7} \\
\midrule
COCA + Optimization & 40.4 & 62.1 & 43.1 & 66.7 & 74.3 & 63.1 & 25.7 & 62.2 & 49.7 & 49.7 & 63.0 & 52.1 \\
\quad + unexplained images & {39.8} & {61.6} & {42.5} & {65.7} & {74.0} & {62.0} & {26.0} & {62.2} & {49.7} & {48.4} & {63.3} & {51.7} \\
\bottomrule
\end{tabular}
}
\caption{Comparison of MLLM-based concept optimization. In the standard setting, we randomly sample images. In the evaluated variant, we specifically select concepts that are not explained by the current concept selection.}
\label{tab:supp:multi-concept:mllmimagesampling}
\end{table}

\subsection{Ablating the Number of Nonzero Concepts in Multi-Level Masking}
\label{sec:supp:multi-target:multilevelmasking}

In \cref{sec:multi-concept}, we propose to select less redundant concepts by multi-level masking, i.e., keeping only the top-10, top-50, and top-100 concept scores per image and setting all others to zero. Here, we ablate whether this method is sensitive to the exact choice of top-10/50/100, and how many masking rounds there are. For both COCA and MLLM-proposed concept pools, we evaluate the multi-level masking strategy for concept selection with 20 rounds of randomly selected top-$k$ triples, where we sample the first $k$ ($k_1$) uniformly in $[5, 20]$, $k_2$ in $[30, 70]$, and $k_3$ in $[80, 120]$. Additionally, we evaluate 20 rounds where we include four masking levels, with top-$k$ values sampled as $k_1$ in $[5, 15]$, $k_2$ in $[20, 40]$, $k_3$ in $[50, 70]$, and $k_4$ in $[80, 120]$. 

Results are in \cref{tab:supp:multi-concept:multilevelmasking}. We find that evaluations with sampled top-$k$ values, including four levels, deviate only minimally from the (10, 50, 100) values used in \cref{tab:complex_data_evaluation_recall} and min/max ranges overall are tight. We conclude that the choice of levels and the top-$k$ values is less important than masking overall.

\begin{table}[h]
\centering
\resizebox{\textwidth}{!}{%
\begin{tabular}{l@{\hspace{8pt}}ccc@{\hspace{8pt}}ccc@{\hspace{8pt}}ccc@{\hspace{8pt}}ccc}
\toprule
& \multicolumn{3}{c}{LVIS} & \multicolumn{3}{c}{MS COCO} & \multicolumn{3}{c}{SUN RGB-D} & \multicolumn{3}{c}{DOTA} \\
\cmidrule(lr){2-4}\cmidrule(lr){5-7}\cmidrule(lr){8-10}\cmidrule(lr){11-13}
{} & MCC-R & Sim. & R$_{\mathrm{J}}$ & MCC-R & Sim. & R$_{\mathrm{J}}$ & MCC-R & Sim. & R$_{\mathrm{J}}$ & MCC-R & Sim. & R$_{\mathrm{J}}$ \\
\midrule
\multicolumn{13}{l}{\textit{MLLM + FG-CLIP + Multi-Level Masking}} \\
\quad Published $(10, 50, 100)$ & 27.5 & 43.1 & 27.5 & 42.4 & 49.4 & 38.2 & 20.6 & 47.7 & 37.3 & 32.4 & 40.0 & 27.6 \\
\quad 3 levels, sampled & 27.4$_{27.0\text{--}28.1}$ & 43.1$_{42.8\text{--}43.3}$ & 27.7$_{26.9\text{--}28.3}$ & 42.1$_{41.4\text{--}42.7}$ & 49.3$_{49.0\text{--}49.8}$ & 38.2$_{37.4\text{--}39.2}$ & 20.6$_{20.0\text{--}21.1}$ & 47.6$_{47.3\text{--}47.8}$ & 36.7$_{36.2\text{--}37.3}$ & 32.5$_{31.7\text{--}34.1}$ & 40.3$_{40.1\text{--}40.6}$ &
  28.0$_{27.2\text{--}29.4}$ \\
\quad 4 levels, sampled & 27.3$_{26.8\text{--}28.0}$ & 43.0$_{42.8\text{--}43.2}$ & 27.6$_{27.0\text{--}27.8}$ & 42.1$_{41.4\text{--}42.5}$ & 49.3$_{49.0\text{--}49.5}$ & 38.0$_{37.4\text{--}38.5}$ & 20.5$_{19.8\text{--}21.2}$ & 47.6$_{47.4\text{--}47.9}$ & 36.8$_{36.2\text{--}37.5}$ & 32.4$_{31.7\text{--}33.8}$ & 40.1$_{39.8\text{--}40.3}$ &
  27.9$_{27.0\text{--}29.7}$ \\
\midrule
\multicolumn{13}{l}{\textit{COCA + FG-CLIP + Multi-Level Masking}} \\
\quad Published $(10, 50, 100)$ & 30.0 & 52.9 & 30.9 & 46.8 & 59.2 & 42.3 & 21.4 & 58.8 & 37.6 & 35.5 & 47.6 & 31.2 \\
\quad 3 levels, sampled & 30.3$_{29.7\text{--}30.9}$ & 53.1$_{52.5\text{--}53.5}$ & 30.8$_{29.3\text{--}32.0}$ & 46.3$_{45.7\text{--}46.9}$ & 59.1$_{58.6\text{--}59.4}$ & 42.0$_{40.6\text{--}42.8}$ & 21.6$_{21.0\text{--}22.2}$ & 59.1$_{58.8\text{--}59.3}$ & 37.6$_{36.8\text{--}38.0}$ & 35.2$_{34.6\text{--}36.0}$ & 47.7$_{47.5\text{--}47.9}$ &
  31.6$_{30.8\text{--}32.4}$ \\
\quad 4 levels, sampled & 30.4$_{29.8\text{--}30.9}$ & 53.1$_{52.7\text{--}53.4}$ & 30.9$_{30.1\text{--}31.9}$ & 46.0$_{45.5\text{--}46.7}$ & 58.9$_{58.3\text{--}59.1}$ & 41.7$_{40.7\text{--}42.4}$ & 21.7$_{21.1\text{--}22.1}$ & 59.1$_{58.9\text{--}59.4}$ & 37.7$_{36.6\text{--}38.8}$ & 35.2$_{34.5\text{--}35.8}$ & 47.6$_{47.4\text{--}47.8}$ &
  31.5$_{30.6\text{--}32.4}$ \\
\bottomrule
\end{tabular}
}
\caption{\textbf{Sensitivity to the masking levels.} Multi-Level Masking with the published per-image top-$K$ levels $(10, 50, 100)$ against 20 randomly sampled level sets per scheme. Sampled rows report the mean over the 20 configurations, with the range (min-max) as a subscript. All other settings (FG-CLIP, Top-$K$ selection, 10 concepts per side) are unchanged.}
\label{tab:supp:multi-concept:multilevelmasking}
\end{table}

\section{Applications: Additionaly Results}
\label{sec:supp:applications:additional-results}

\subsection{Additional ImageNet Spurious Correlations}
\label{sec:supp:applications:additional-results:imagenet}

\begin{figure}[p]
\centering
\includegraphics[width=0.9\linewidth]{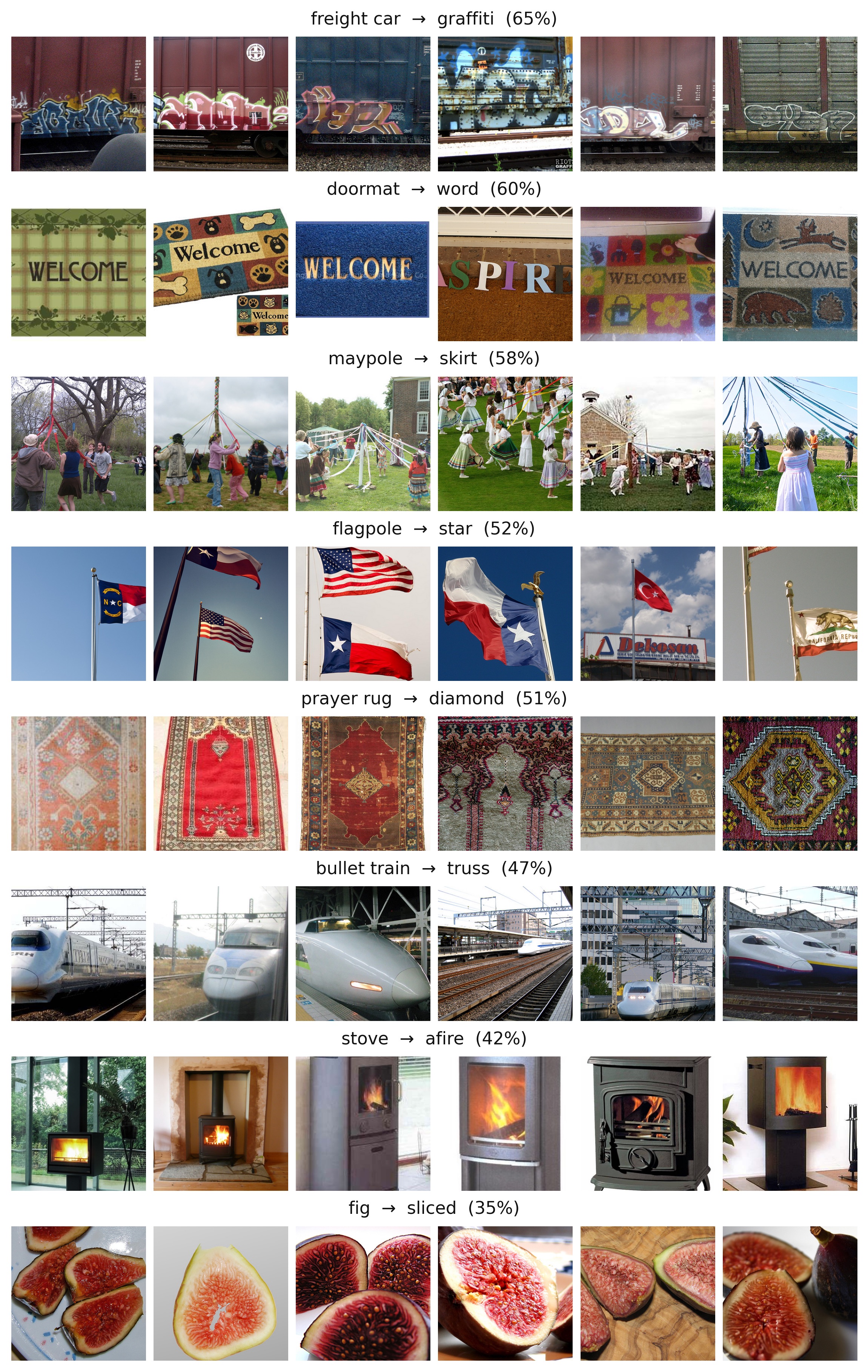}
\caption{Examples of discovered class–concept correlations in ImageNet. Each row displays representative images for a class alongside the correlated concept and its MLLM-verified presence score (in parentheses).}
\label{fig:supp:imagenet-correlations-additional}
\end{figure}

In \cref{sec:applications}, we use FG-CLIP embeddings and dataset diffing to find candidates for spurious class-concept correlations in ImageNet. Of the 6,103 proposed candidates, we verified 1,553 class-concept pairs with an MLLM. This still contains a mix of genuinely interesting correlations, obvious or expected associations, and borderline and ambiguous cases. Therefore, the final list requires manual inspection. Furthermore, many correlations are specific to one class and one concept and cannot be generalized, which makes a systematic presentation impossible. Instead, we selected eight of the discovered correlations that we found interesting in \cref{fig:supp:imagenet-correlations-additional}.

Specifically, we found a large number of freight trains with graffiti on them (65\% MLLM presence score), which aligns with the \enquote{decay} theme of objects. We found that many doormats have words on them (60\% presence score), some \enquote{maypole} images show traditional female dresses (58\%), and many prayer rugs show a diamond or route pattern (51\%), which has symbolic meaning in Muslim cultures. One very interesting example is the high number of stars on flags in the \enquote{flagpole} class. This is not exclusive to US \enquote{stars-and-stripes} flags, but also includes, for example, the Turkish, Californian, and Texan flags. Finally, we found many bullet train images with steel constructions (trusses), which is perhaps less surprising, but also many ovens with fire burning and sliced figs, which are peculiar distributions.

\subsection{Additional Results for Demographic Bias Detection in T2I Models}
\label{sec:supp:applications:demographicbias}

\begin{figure}[p]
\centering
\begin{subfigure}[b]{\textwidth}
\centering
\includegraphics[width=\textwidth]{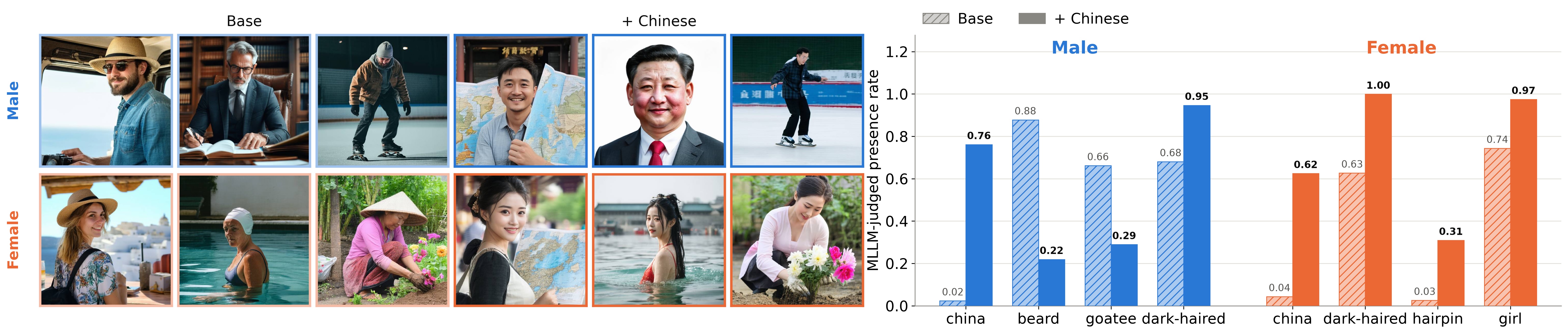}
\caption{Example detected biased attributes for \enquote{Chinese} (country) demographic group.}
\label{fig:supp:applications:demographicbias:chinese}
\end{subfigure}
\vfill
\begin{subfigure}[b]{\textwidth}
\centering
\includegraphics[width=\textwidth]{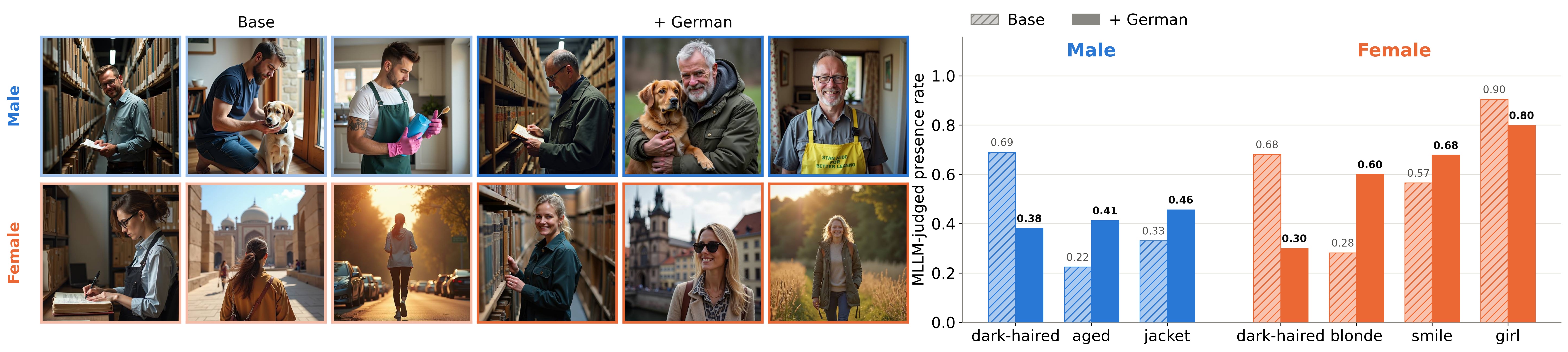}
\caption{Example detected biased attributes for \enquote{German} (country) demographic group.}
\label{fig:supp:applications:demographicbias:german}
\end{subfigure}
\vfill
\begin{subfigure}[b]{\textwidth}
\centering
\includegraphics[width=\textwidth]{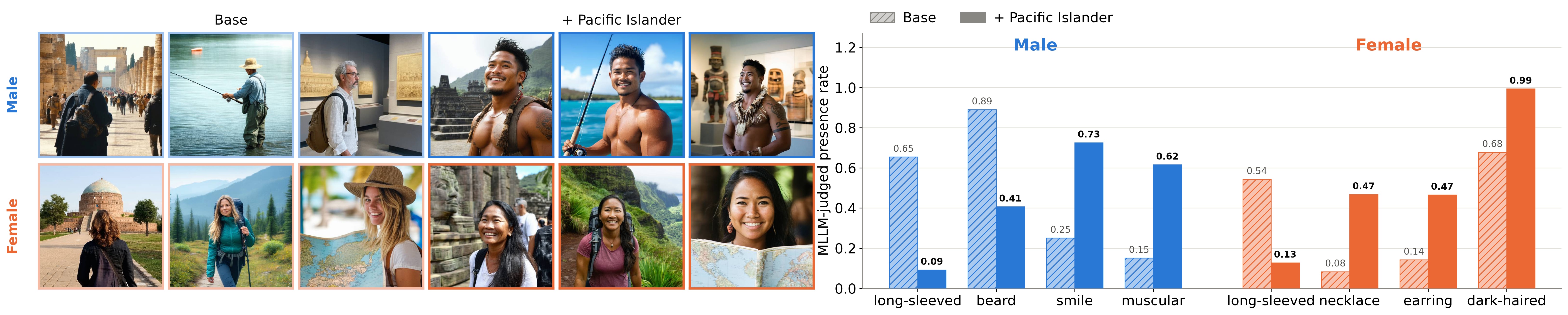}
\caption{Example detected biased attributes for \enquote{Pacific Islander} (race) demographic group.}
\label{fig:supp:applications:demographicbias:pacificislander}
\end{subfigure}
\vfill
\begin{subfigure}[b]{\textwidth}
\centering
\includegraphics[width=\textwidth]{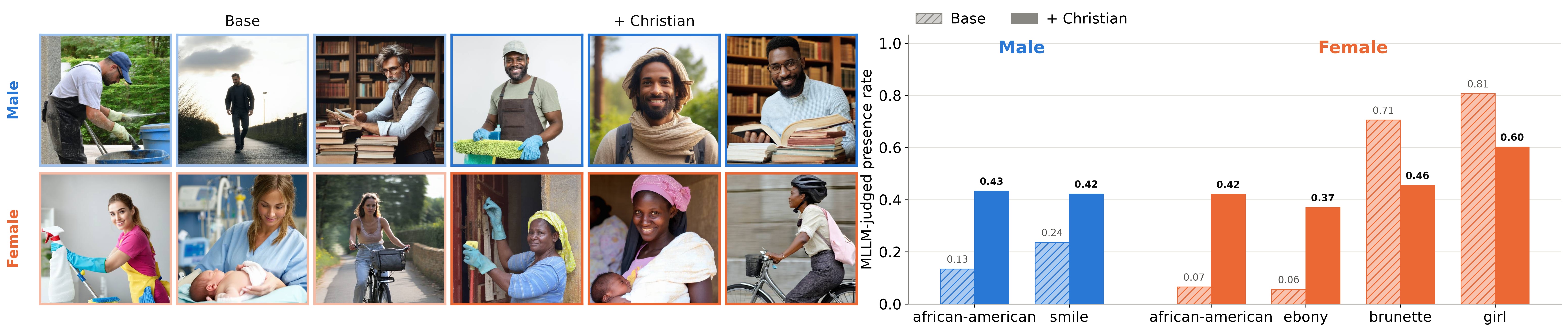}
\caption{Example detected biased attributes for \enquote{Christian} (religion) demographic group.}
\label{fig:supp:applications:demographicbias:christian}
\end{subfigure}
\vfill
\begin{subfigure}[b]{\textwidth}
\centering
\includegraphics[width=\textwidth]{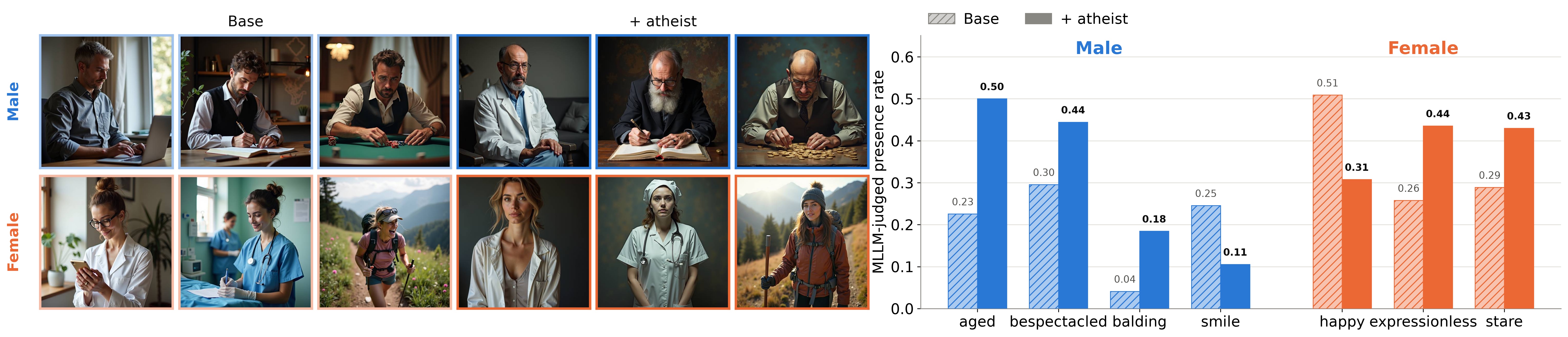}
\caption{Example detected biased attributes for \enquote{atheist} (religion) demographic group.}
\label{fig:supp:applications:demographicbias:atheist}
\end{subfigure}
\vfill
\begin{subfigure}[b]{\textwidth}
\centering
\includegraphics[width=\textwidth]{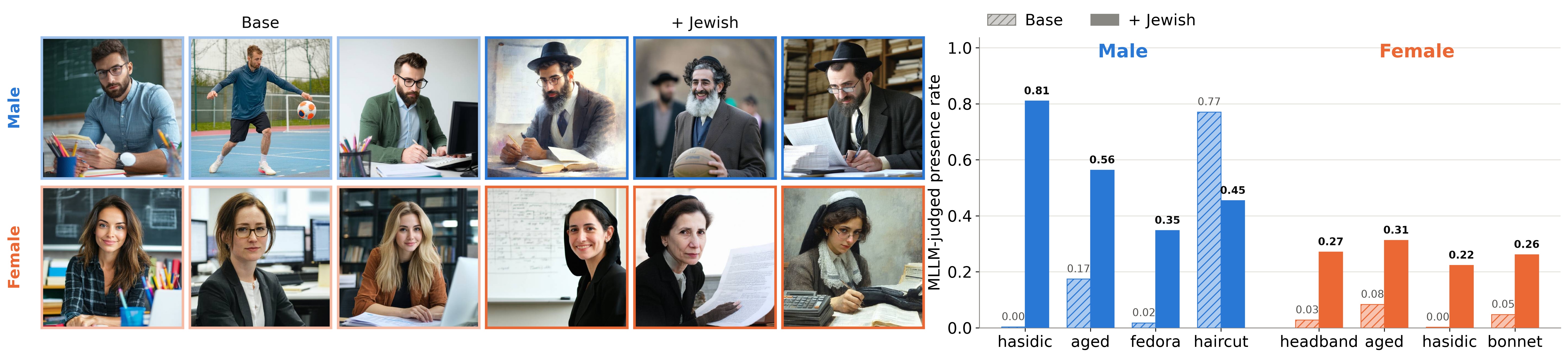}
\caption{Example detected biased attributes for \enquote{Jewish} (religion) demographic group.}
\label{fig:supp:applications:demographicbias:jewish}
\end{subfigure}
\caption{Examples of detected biased attributes across country/religion/race demographic groups across Flux and SD-3.5-Large T2I models.}
\label{fig:supp:applications:demographicbias}
\end{figure}

In \cref{fig:supp:applications:demographicbias}, we show additional examples of highly significant demographic biases in images generated by Flux and SD-3.5-Large. For countries, we show examples for \enquote{Chinese} (\cref{fig:supp:applications:demographicbias:chinese}) and \enquote{German} (\cref{fig:supp:applications:demographicbias:german}). We find that, relative to the base prompt, Chinese men are portrayed with dark hair and less facial hair, while German men are portrayed older and less with dark hair. For women, we find that the same haircolor and age pattern holds. Another interesting finding is specific accessories, such as \enquote{jacket} for German men and \enquote{hairpin} for Chinese women. The distributional differences may be induced by genuinely different demographics, cultural stereotypes, or skewed sampling of the training data. Assessing which biases are harmful or undesired is outside of the scope of this work.

Another interesting example is \enquote{Pacific Islander} as a race group (\cref{fig:supp:applications:demographicbias:pacificislander}). Both men and women are portrayed as showing more bare skin (instantiated by the reduced \enquote{long-sleeved} attribute), and Pacific Islander women are portrayed as wearing more earrings and necklaces. The increased prevalence of jewelry for women, together with the increase of \enquote{smiling} for both men and women, is a recurring pattern for demographic groups, especially nonwestern groups.

Finally, we compare outputs for religious groups. We find that all religions except \enquote{Christian} are portrayed in a very marked way. For example, \enquote{Buddhist} persons are almost exclusively displayed as monks or nuns. \enquote{Jewish} prompts generate mostly images where people look \enquote{Hasidic} (\cref{fig:supp:applications:demographicbias:jewish}), e.g., men wearing hats (\enquote{fedora}) and women’s bonnets. Both genders overall show less hair and are portrayed older. Across most demographic groups, age shifts are among the most significant shifts. Images for \enquote{Christian} (\cref{fig:supp:applications:demographicbias:christian}) overall differ less from the base distribution, but include more dark-skinned people (\enquote{African-American}, \enquote{ebony}) and women appear older (20\% less prevalence of \enquote{girl}). Finally, we noticed that \enquote{atheist} ( \cref{fig:supp:applications:demographicbias:atheist}) images also differ from the base distribution in a peculiar way: Men appear older, more balding, and wearing more glasses. Women appear \enquote{expressionless} and directly face the viewer (\enquote{stare}). Both groups are portrayed less happy.

\section{Detailed Description of Methods}
\label{sec:supp:methods}

\subsection{Propose-then-Rank}
\label{sec:supp:methods:propose-then-rank}

The propose-then-rank baseline follows \citet{dunlap2024describing} and consists of two stages: (i) MLLM-driven candidate generation from visual prompts, and (ii) embedding-based ranking over the full image sets.

\mypara{Candidate Generation (\enquote{Propose})}
Let $k$ denote the number of proposal iterations (we use $k=5$). In each iteration $r \in \{1, \ldots, k\}$, we randomly sample five images from set $A$ and five images from set $B$. These images are arranged in a $2 \times 5$ grid, with set $A$ in the top row and set $B$ in the bottom row. The grid is presented to an MLLM together with a prompt requesting $n$ short visual concepts that characterize set $A$ relative to set $B$, and $n$ concepts that characterize set $B$ relative to set $A$ (we use $n=10$). The MLLM returns a structured list of candidate phrases, such as \enquote{outdoor scene} or \enquote{bright lighting}. After $k$ iterations, we collect all proposed candidates and deduplicate them to obtain a vocabulary $\mathcal{S}$ of unique concepts.

\mypara{Ranking}
For ranking, we evaluate every candidate $s \in \mathcal{S}$ against all images in both sets using image-text similarity. Let $\phi(x)$ denote the normalized image embedding of image $x$ and $\psi(s)$ the normalized text embedding of concept $s$, both obtained from SigLIP2. We calculate $f_{\mathrm{emb}}(s, x)$ (see \cref{sec:method}).\footnote{For SigLIP models, we additionally apply sigmoid: $f(s, x) = \sigma\!\left(\exp(\ell) \cdot \phi(x)^\top \psi(s) + b\right)$, where $\sigma$ is the sigmoid function, and $\ell$ and $b$ are the learned logit scale and bias parameters of SigLIP2.}

For each candidate $s$, we compute its directional discriminativeness via the area under the ROC curve (AUROC), following \citet{dunlap2024describing}. We find that when evaluating a large number of potentially irrelevant concepts, AUROC is unsuitable, because it admits concepts that have generally low similarity, but spuriously separate the sets. However, in the case of propose-then-rank, all concepts are relevant to the images because the MLLM proposes them based on the seen images, so the likelihood of introducing noise is low. We denote the AUROC score of a candidate $s$ as $\operatorname{AUC}_A(s)$ when treating $A$ as the positive class, and $\operatorname{AUC}_B(s)$ when treating $B$ as the positive class. Then, we select the top-$N$ candidates by $\operatorname{AUC}_A$ as the final feature set $\widehat{\mathcal{S}}_A$ for set $A$, and the top-$N$ candidates by $\operatorname{AUC}_B$ as $\widehat{\mathcal{S}}_B$ for set $B$ (we use $N=10$ unless stated otherwise).

\subsection{Distribution Shift Explanations}
\label{sec:supp:methods:distribution-shift}

The distribution shift baselines aim to discover the semantic differences between sets $A$ and $B$ by first identifying geometric directions in the image embedding space that capture the shift, and then decoding these directions into natural language concepts. This process consists of two stages: (i) subspace discovery in the joint embedding space, and (ii) vocabulary mapping and aggregation.

Let $\phi(x), \phi(y) \in \mathbb{R}^D$ denote the L2-normalized SigLIP2 image embeddings for images $x \in A$ and $y \in B$. We seek a set of orthogonal directions $u_c \in \mathbb{R}^D$ that best separate the two distributions. We evaluate two methods for finding these directions: Linear Support Vector Classification (LinearSVC) and U-WaX \citep{naumann2026wasserstein}.

\mypara{Subspace Discovery}
For the \textbf{Linear} approach, we iteratively discover a set of orthogonal directions. In each step $c$, we train a LinearSVC (with $A$ as the positive class and $B$ as the negative class) on the current embeddings to obtain a weight vector $w_c$. We orthogonalize $w_c$ against all previously found directions using Gram-Schmidt orthogonalization, and L2-normalize it to obtain $u_c$. We then project all embeddings onto the orthogonal complement of $u_c$, i.e., $\phi(x) \leftarrow \phi(x) - (u_c^\top \phi(x))u_c$, and repeat the process to find the next direction.

For the \textbf{U-WaX} approach, we frame the shift as an optimal transport problem. We compute the pairwise squared L2 cost matrix between all $\phi(x)$ and $\phi(y)$, and solve for the optimal unregularized transport plan $\gamma^\ast$ that couples instances from $A$ and $B$. U-WaX then extracts orthogonal directions $u_c$ that maximize the Wasserstein distance attributed to these specific subspaces, effectively isolating the sub-shifts that contribute most to the structural displacement between the sets.

\mypara{Vocabulary Mapping and Aggregation}
Once the shift directions $\{u_c\}$ are identified, we map them back to interpretable text concepts. Given a large vocabulary of candidate concepts $\mathcal{S}$ and their L2-normalized text embeddings $\psi(s)$, we compute the projection of each concept onto the shift direction via $\psi(s)^\top u_c$. We use the same list of 54,000 COCA terms introduced in \cref{sec:evaluation}.

Because the directions are oriented such that positive values indicate alignment with set $A$ and negative values indicate alignment with set $B$, concepts with the highest positive scores are retrieved as candidate features for $A$. Conversely, concepts with the lowest negative scores (i.e., highest alignment with $-u_c$) are retrieved as candidate features for $B$.

\subsection{Tagging and Aggregation Methods}
\label{sec:supp:methods:tagging}

The tagging and aggregation baselines operate by extracting textual representations from each image individually and subsequently summarizing these representations using an LLM. This approach aims to bypass the direct evaluation of image embeddings by mapping images to text first, thereby reframing the dataset difference problem as a text summarization task over discrete image attributes. 

These baselines consist of two distinct stages: (i) per-image tagging, where textual features or captions are extracted for all images $x \in A$ and $y \in B$, and (ii) aggregation, where the collected textual data is summarized to identify the distinguishing concepts between the sets.

\mypara{Per-Image Tagging}
We evaluate three variants for extracting textual representations from the images, differing in the format and the model used:
\begin{enumerate}
    \item \textbf{Caption Summary:} For each image, we use an MLLM (Gemma4-31B) to generate a detailed, long-form caption. This provides a holistic but unstructured description of the image content.
    \item \textbf{Image Tagging (MLLM):} We prompt the MLLM to return a structured JSON output containing a list of all identified objects and their visual attributes (e.g., color, texture, shape, state). This potentially isolates atomic concepts better than free-form captions.
    \item \textbf{Image Tagging (RAM++):} We use the RAM++ model \citep{zhang2024recognize} to extract a set of predefined visual concepts (tags) present in each image. Because RAM++ operates on a fixed vocabulary, its outputs are intrinsically normalized across the dataset.
\end{enumerate}

\mypara{Aggregation}
For the \textbf{Caption Summary} and \textbf{Image Tagging (MLLM)} variants, the extracted texts are highly variable. We address this by partitioning the captions into smaller subsets and processing them via repeated queries to an LLM. Specifically, we randomly sample small batches of captions from set $A$ and set $B$. For each batch, we prompt an LLM (DeepSeek V4 Flash) to infer the semantic theme that distinguishes the two sets and to propose 10 characteristic features for $A$ and 10 for $B$. This process is repeated across multiple independent queries to ensure broad coverage. Finally, all the proposed feature lists are concatenated and passed to the LLM in a consolidation prompt, which synthesizes the repeated observations into a final, representative list of 10 distinguishing features for set $A$ and 10 for set $B$.

For the \textbf{Image Tagging (RAM++)} variant, the aggregation is more straightforward due to the fixed vocabulary. Let $\mathcal{T}$ denote the set of all unique tags identified across both datasets. For each tag $t \in \mathcal{T}$, we compute its frequency in set $A$ and set $B$, denoted as $p_A(t)$ and $p_B(t)$. We then define the margin for a tag as the difference in proportions: $\delta(t) = p_A(t) - p_B(t)$. The top 10 tags with the highest positive margins are selected as the characteristic features for set $A$, and the top 10 tags with the lowest negative margins are selected for set $B$.

\subsection{Wordlist (COCA) + SigLIP or FG-CLIP Embeddings}
\label{sec:supp:methods:fg-clip}

In this paper, we use SigLIP2 image embeddings and FG-CLIP patch-level embeddings to discover visual concepts that distinguish sets $A$ and $B$. FG-CLIP operates on localized image patches, while SigLIP operates on global representations, so FG-CLIP can capture small objects and attributes that might be overlooked by whole-image embeddings. In both cases, the method includes two steps: (i) image embedding and concept scoring, and (ii) vocabulary ranking based on concept score margins.

\mypara{Image Embedding and Concept Scoring}
Given images $x \in A$ and $y \in B$, we use SigLIP2 \citep{tschannen2025siglip} or FG-CLIP2 \citep{xie2025fg2} to extract image embeddings. In the case of SigLIP, the image encoder directly yields a single embedding vector $\phi(x)$. FG-CLIP instead returns a matrix of patch embeddings for each image. Here, $\phi(x) \in \mathbb{R}^{D \times \rho}$ is the matrix of L2-normalized patch embeddings for image $x$, where $\rho$ is the number of patches and $\phi_i(x) \in \mathbb{R}^D$ is the embedding of the $i$-th patch. 

We evaluate these images against a large pool of candidate text concepts $\mathcal{C}$, for which we compute L2-normalized text embeddings $\psi(s) \in \mathbb{R}^D$ using the SigLIP or FG-CLIP text encoder. As with the other methods, we use a vocabulary derived from the COCA corpus containing $M = 54,000$ concepts. To determine the relevance of a concept $s$ to an image $x$, we compute the similarity score $f_{\mathrm{emb}}(s, x)$. For SigLIP, this is calculated as
\begin{equation}
    f_{\mathrm{emb}}(s, x) = \sigma(\exp(\ell) \cdot \psi(s)^T \phi(x) + b),
\end{equation}
where $\sigma$ is the sigmoid function, and $\ell$ and $b$ are the learned logit scale and bias parameters of the SigLIP model. For FG-CLIP, we compute the max-over-patches cosine similarity:
\begin{equation}
    f_{\mathrm{emb}}(s, x) = \max_{i \in \{1, \ldots, \rho\}} \psi(s)^\top \phi_i(x).
\end{equation}
This formulation ensures that a concept is scored highly if it aligns strongly with any local region in the image. We compute $f_{\mathrm{emb}}$ for all candidate concepts across all images in both $A$ and $B$.

\mypara{Vocabulary Ranking}
After scoring all images against the vocabulary, we aggregate the scores to find the concepts that best distinguish the two datasets. For each concept $s$, we compute a divergence $\hat{\delta}_{A, B}(s)$ between the score distribution for set $A$ and the score distribution for set $B$. This divergence can take multiple forms. In \cref{sec:evaluation}, we use the difference of means:
\begin{align}
    \hat{\delta}_{A,B}(s) &=
    \mathbb{E}_{x\sim A}[f_{\mathrm{emb}}(s, x)] - \mathbb{E}_{y\sim B}[f_{\mathrm{emb}}(s, y)] \\
    &=
    \frac{1}{|A|}\sum_{x \in A} f_{\mathrm{emb}}(s, x) -
    \frac{1}{|B|}\sum_{y \in B} f_{\mathrm{emb}}(s, y),
\end{align}
which is easy to calculate and effective. In \cref{sec:multi-concept}, we additionally divide the difference of means by the standard deviation of the concept's scores over both sets, i.e., we use a standardized effect size:
  \begin{align}
      \overline{f}(s) &= \frac{1}{|A|+|B|}\sum_{x \in A \cup B} f_{\mathrm{emb}}(s, x) \\
      \sigma(s) &= \sqrt{\frac{1}{|A|+|B|} \sum_{x\in A \cup B} \left(f_{\mathrm{emb}}(s, x) - \overline{f}(s) \right)^2}\\
      \hat{\delta}_{A,B}(s) &= \frac{1}{\sigma(s)}\left(\frac{1}{|A|}\sum_{x \in A} f_{\mathrm{emb}}(s, x) -
      \frac{1}{|B|}\sum_{y \in B} f_{\mathrm{emb}}(s, y)\right).
  \end{align}
This makes concepts whose scores have very different spreads comparable.

In \cref{sec:applications}, we also use $\operatorname{AUROC}$ as divergence:
\begin{equation}
    \begin{aligned}
        \hat{\delta}_{A,B}(s) = \frac{1}{|A||B|} \sum_{x \in A} \sum_{y \in B} \Bigg[ & \mathbb{I}\left( f_{\mathrm{emb}}(s, x) > f_{\mathrm{emb}}(s, y) \right) \\
        & + \frac{1}{2} \mathbb{I}\left( f_{\mathrm{emb}}(s, x) = f_{\mathrm{emb}}(s, y) \right) \Bigg],
    \end{aligned}
\end{equation}
where $\mathbb{I}(\cdot)$ is the indicator function. The $\operatorname{AUROC}$ estimates the probability that a randomly chosen image from set $A$ receives a higher concept score than a randomly chosen image from set $B$.

We then rank the concepts based on the score margin $\hat{\delta}_{A,B}(s)$. The concepts with the largest positive margins represent features strongly associated with set $A$ relative to $B$, while those with the largest negative margins (i.e., highest $\hat{\delta}_{B,A}(s)$) represent features characteristic of set $B$.

\mypara{MLLM Optimization}
To discover specific visual differences that are not captured in the static wordlist, such as rare concepts and multiword expressions, we optionally perform an iterative refinement of the candidate concepts using an MLLM. We initialize the optimization with the top-ranked concepts (by score margin $\hat{\delta}_{A,B}(s)$) from the wordlist. 

\begin{figure}[t]
\centering
\includegraphics[width=\linewidth]{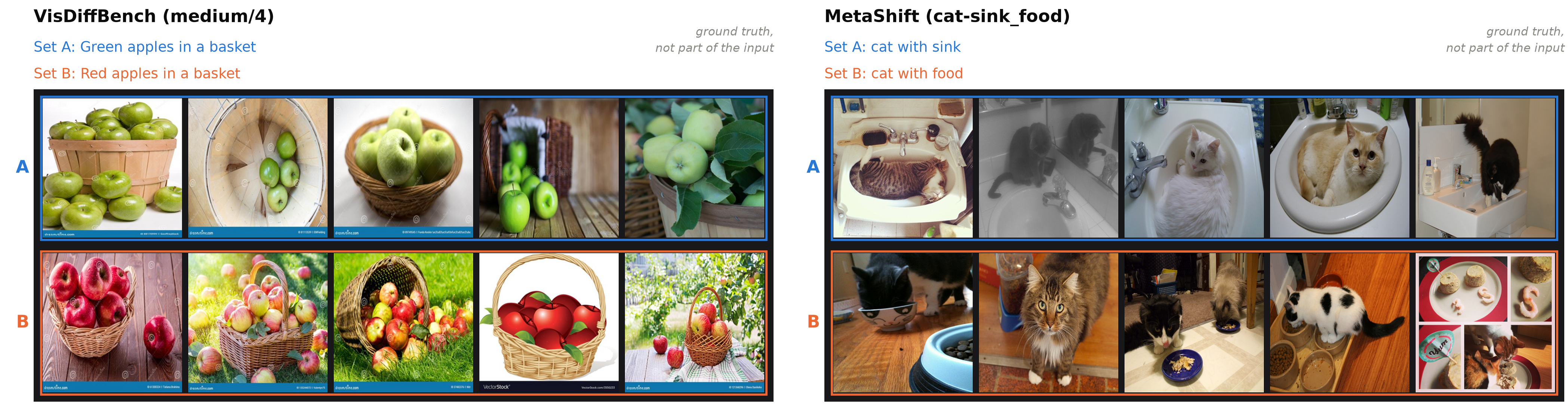}
\caption{Example visual input grids provided to the MLLM during optimization. Each grid displays sample images from Set $A$ (top row, blue) and Set $B$ (bottom row, orange) for tasks from VisDiffBench (left) and MetaShift (right). Ground-truth descriptions above each grid are shown for reference only and are not included in the model's input.}
\label{fig:supp:mllm-optimization-input}
\end{figure}

At each iteration, the MLLM is provided with an image grid containing examples from both set $A$ and set $B$, see \cref{fig:supp:mllm-optimization-input} for an example. Alongside this visual context, the MLLM is given a historical record of currently active and previously discarded textual concepts, along with their discriminative metrics (such as the difference of means and $\operatorname{AUROC}$).

Prompted with this multimodal context, the MLLM infers the underlying semantic themes and proposes novel, semantically distinct candidate phrases to better separate the two sets. We then compute the text embeddings $\psi(s)$ for these new proposals, evaluate their similarity scores $f_{\mathrm{emb}}(s, x)$ against the image sets, and calculate their score margins $\hat{\delta}_{A,B}(s)$. After filtering out semantically redundant phrases, the newly proposed concepts are pooled with the existing candidates, and we retain only the top-$k$ most discriminative concepts for the next iteration.

\mypara{Combining SigLIP and FG-CLIP Scores}
To use the complementary strengths of global and local representations, we combine the scores from both SigLIP and FG-CLIP to rank concepts. Because the raw similarity scores $f_{\mathrm{emb}}(s, x)$ from the two models are on different scales and follow different distributions, we first map them to a common scale before comparison. To achieve this, we calibrate the cosine similarities of each model to presence probabilities $p(s, x) = \sigma\left(a \cdot \psi(s)^\top \phi(x) + b\right)$, where $a$ and $b$ are fitted once per model by logistic regression on the object-presence labels of SUN RGB-D (Platt scaling). For FG-CLIP, $\psi(s)^\top \phi(x)$ is the max-over-patches similarity. After calibration, we calculate the divergence $\hat{\delta}_{A,B}(s)$ as the difference of mean presence probabilities in both SigLIP space and FG-CLIP space, which are now directly comparable. Then, we take the maximum divergence score over the two spaces, i.e., formally, the combined divergence is defined as
\begin{equation}
    \hat{\delta}_{A,B}(s) = \max\left(\hat{\delta}_{A,B}^{\mathrm{SigLIP}}(s), \hat{\delta}_{A,B}^{\mathrm{FG\text{-}CLIP}}(s)\right).
\end{equation}
By taking the maximum, a concept is deemed highly discriminative if it effectively separates the two image sets in either the global representation or the fine-grained view.

\section{Technical Details of the Multi-Target Evaluation}
\label{sec:supp:multi-target}

\subsection{How we Sample the Multi-Target Datasets}
\label{sec:supp:multi-target:sampling}

To simulate complex distributions with overlapping concepts, we cannot sample images for each concept independently because multi-object co-occurrence (in one image) entails that assigning an image to a set to satisfy the demand for one concept will at the same time change the counts of all other concepts present in that image. To address this challenge, we frame multi-target dataset construction as an optimization problem over the assignments of all images.

\mypara{Target Definition} 
Let $n$ be the target size for each set ($A$ and $B$). For every concept $c$ in the dataset (e.g., COCO or SUN RGB-D classes), we assign a role (aligned to $A$, aligned to $B$, or neutral), an alignment ratio $r_c$, and an overall prevalence $p_c$. The overall prevalence is identical to the prevalence of the concept in the source dataset (e.g., COCO, LVIS, or SUN RGB-D), with a random perturbation that allows up to $\pm 10\%$ deviation from the original prevalence. From the roles, alignment ratios $r_c$, and prevalence $r_c$, we compute count targets $t^A_c$ and $t^B_c$ for Set $A$ and Set $B$. For example, if a concept is aligned to Set $A$, its count target satisfies $\frac{t^A_c}{t^A_c + t^B_c} \ge r_c$.

\mypara{Objective Function}
We define a state where each image $i$ is assigned to Set $A$, Set $B$, or excluded. Let $c^A_c$ and $c^B_c$ be the realized counts of concept $c$ in the current assignment, and let $|A|$ and $|B|$ be the current sizes of the sets. We minimize a loss function $\mathcal{L}$ that penalizes squared deviations from both the concept targets and the overall set size limits:
\begin{equation}
\mathcal{L} = \sum_{c} \left[ (c^A_c - t^A_c)^2 + (c^B_c - t^B_c)^2 \right] + \lambda \left[ (|A| - n)^2 + (|B| - n)^2 \right]
\end{equation}
where $\lambda=1.0$ balances the concept counts with the set size constraints.

\mypara{Initialization (Greedy Assignment)}
We greedily initialize the sets $A$, $B$, and excluded images. For each image, we calculate the unmet demand in Set $A$ and Set $B$ for the concepts it contains. The image is assigned to the set with the highest unmet demand, provided that set has not yet reached the maximum capacity $n$. If neither set needs the image's concepts, it is excluded.

\mypara{Iterated Local Search}
The greedy initialization provides a strong starting point but is sub-optimal due to complex co-occurrences which cannot be satisfied greedily. To refine the sets, we apply an iterated local search: In each step, we randomly select an image and evaluate the loss $\mathcal{L}$ for all three of its possible states (assigned to $A$, assigned to $B$, or excluded), keeping the assignments of all other images fixed. The image is reassigned to the state that strictly minimizes the global loss. This process trades off local shortfalls against competing per-class targets and is repeated until convergence (i.e., when no improvements are found after 20,000 iterations).

\subsection{Multi-Level Masked Concept Selection}
\label{sec:supp:multi-target:maskedselection}

Here, we explain how we use multi-level masked concept selection to select more diverse concepts from the overall pool (MLLM-proposed or from a static wordlist).
Let $\mathbf{S} \in \mathbb{R}^{(|A|+|B|) \times P}$ be the score matrix evaluated over all images in sets $A$ and $B$, where the rows are indexed by the images and the columns are indexed by the $P$ candidates in the pool (e.g., the COCA word list). Let $\mathbf{y} \in \{0,1\}^{|A|+|B|}$ indicate set membership (in $A$ or in $B$).

\mypara{Per-Image Top-$K$ Masking}
For a given threshold level $K$, we define the masked score matrix $\mathbf{S}^{(K)}$ by retaining, in each row, only the $K$ highest-scoring candidates for that specific image:
\begin{equation}
\mathbf{S}^{(K)}_{i, j} = 
\begin{cases} 
\mathbf{S}_{i, j} & \text{if } j \in \operatorname{top}_K(\mathbf{S}_{i, :}), \\
0 & \text{otherwise},
\end{cases}
\end{equation}
where $\operatorname{top}_k(\mathbf{S}_{i, :})$ is the index set of the $k$ largest entries in the $i$-th row. This procedure suppresses candidates that are diffusely elevated across many images without ever being a strong match for any single image, while preserving genuine concepts that are strongly present in the image.

\mypara{Per-Level Selection}
For each masking level $K \in \mathcal{K} = \{10, 50, 100\}$, we compute the score divergence from the masked matrix:
\begin{equation}
\delta^{(K)}{(s_j)} = \operatorname{sign} \cdot \left( \frac{1}{|A|}\sum_{i \in A} \mathbf{S}^{(K)}_{i, j} - \frac{1}{|B|}\sum_{i \in B} \mathbf{S}^{(K)}_{i, j} \right).
\end{equation}
Here, we use the standardized image-concept similarity scores. We then apply a selection method $g$ (such as Top-$k$ or Residual-OMP) to the masked matrix and divergence scores and get an ordered list of 10 candidates:
\begin{equation}
L^{(K)} = f(\mathbf{S}^{(K)}, \delta^{(K)}) = \big(c^{(K)}_1, \dots, c^{(K)}_{10}\big).
\end{equation}
We run this independently for each $K \in \mathcal{K}$, which yields three rank-ordered lists $L^{(10)}, L^{(50)}$, and $L^{(100)}$ per target side ($A$ and $B$).

\mypara{Round-Robin Aggregation}
Finally, the three candidate lists are merged into a single list by round-robin interleaving over ascending $K$ and rank $r$:
\begin{equation}
\big(c^{(10)}_1, c^{(50)}_1, c^{(100)}_1, c^{(10)}_2, c^{(50)}_2, c^{(100)}_2, \dots \big).
\end{equation}
This combined list is deduplicated by keeping only the first occurrence of each concept and then truncated to 10 concepts, so a candidate concept that is re-selected at a later level does not consume an additional slot, and we return exactly 10 concepts.

\subsection{Residual Orthogonal Matching Pursuit (Residual-OMP)}
\label{sec:supp:multi-target:residual-omp}

In \cref{sec:multi-concept}, we find that in complex and noisy datasets, naive top-$k$ selection of concepts based on the score divergence is suboptimal, because the top-ranked concepts are often redundant or repetitive. Therefore, we propose and evaluate orthogonal matching pursuit (OMP) as a method to select a subset of $k$ high-scoring concepts that is also diverse. Concretely, the \textbf{Residual-OMP} (Orthogonal Matching Pursuit) method selects a subset of concepts by greedily capturing the variance in the target binary labels that remains unexplained by previously selected concepts.

Let $\mathbf{S} \in \mathbb{R}^{(|A|+|B|) \times M}$ be the matrix of concept scores evaluated over all images in sets $A$ and $B$, where $S_{i, j} = f_{\mathrm{emb}}(s_j, x_i)$. We first standardize the columns (concepts) of $\mathbf{S}$ such that each concept has zero mean and unit variance across the dataset, yielding the normalized design matrix $\mathbf{X}$. Let $\mathbf{y} \in \{0, 1\}^{|A|+|B|}$ be the binary indicator vector where $y_i = 1$ if $x_i \in A$ and $0$ otherwise.

At the beginning of the selection process, we initialize the residual target vector as $\mathbf{r}^{(0)} = \mathbf{y}$ and the set of selected concept indices as $\mathcal{K} = \emptyset$. At each iteration $t \in \{1, \ldots, k\}$ (where $k$ is the target budget), we select the concept $j^\ast \notin \mathcal{K}$ that maximizes the absolute correlation with the current residual:
\begin{equation}
    j^\ast = \arg\max_{j \notin \mathcal{K}} \left| \mathbf{X}_{:, j}^\top \mathbf{r}^{(t-1)} \right|.
\end{equation}
We then update the active set $\mathcal{K} \leftarrow \mathcal{K} \cup \{j^\ast\}$. Next, we orthogonalize the target vector against the subspace spanned by all selected concepts. Let $\mathbf{X}_{\mathcal{K}}$ be the submatrix of $\mathbf{X}$ containing only the columns in $\mathcal{K}$. We fit a linear model to predict $\mathbf{y}$ using $\mathbf{X}_{\mathcal{K}}$ (via least squares) and compute the new residual:
\begin{equation}
    \mathbf{r}^{(t)} = \mathbf{y} - \mathbf{X}_{\mathcal{K}} (\mathbf{X}_{\mathcal{K}}^\top \mathbf{X}_{\mathcal{K}})^{-1} \mathbf{X}_{\mathcal{K}}^\top \mathbf{y}.
\end{equation}
This process is repeated until $k$ concepts are selected. By refitting the linear model on the full active set and updating the residual at each step, Residual-OMP naturally penalizes redundancy, selecting concepts that provide orthogonal discriminative signal rather than merely restating information captured by earlier selections.

\subsection{Detailed Description of Metrics for Multi-Target Evaluation}
\label{sec:supp:multi-concept:metrics}

In \cref{sec:multi-concept}, we evaluate methods using three metrics that focus on ground-truth concept recall from different perspectives:

\textbf{MCC-R} measures whether the selected concepts, together with their similarity scores $f_{\mathrm{emb}}$, can predict the binary ground-truth concept membership. An aligned ground-truth concept $c$ counts as recovered if and only if one of the $K$ predicted concepts reaches a Matthews Correlation Coefficient (MCC) of at least $\tau=0.5$ against the binary class membership labels, for any threshold on the continuous $f_{\mathrm{emb}}$ image-concept scores, i.e., for each concept we take the maximum over all thresholds $\max_{t} \mathrm{MCC}\big(\mathbf{1}[f_{\mathrm{emb}(s)} \ge t],\ y_c\big)$ where $y_c$ are the binary class memberships. For each task, we calculate the ratio of ground truth concepts that can be recovered in this way as a recall metric.

In general, the MCC for a specific binarization threshold $t$ is defined as:
\begin{equation}
    \operatorname{MCC}(t) = 
    \frac{\operatorname{tp}(t)\operatorname{tn}(t) - \operatorname{fp}(t)\operatorname{fn}(t)}{\sqrt{\left(\operatorname{tp}(t) + \operatorname{fp}(t)\right)\left(\operatorname{tp}(t) + \operatorname{fn}(t)\right)\left(\operatorname{tn}(t) + \operatorname{fp}(t)\right)\left(\operatorname{tn(t) + \operatorname{fn(t)}}\right)}},
\end{equation}
where $\operatorname{tp}(t)$ are the true positives when evaluating the ground-truth per-image class memberships against the binarized score memberships, and equivalently for $\operatorname{tn}$ (true negatives), $\operatorname{fp}$ (false positives), $\operatorname{fn}$ (false negatives). Intuitively, MCC quantifies the overall correlation between the induced binary classifications and the ground-truth binary labels across all four quadrants of the confusion matrix ($\operatorname{tp}$, $\operatorname{tn}$, $\operatorname{fp}$, $\operatorname{fn}$), which provides a balanced score even when the classes are heavily imbalanced. This is the case for many aligned classes in our setup that have low overall prevalence.

We use MCC-R to provide a holistic understanding of how well a method, which includes the concept selection and the image-concept scores, can reconstruct the structure of the dataset, i.e., the aligned concepts and their per-image presences. This is a realistic assessment because in deployment, a user would only have access to the predicted concepts and their scores, which must be used to reason about the data.

\begin{figure}[h]
\centering
\includegraphics[width=\textwidth]{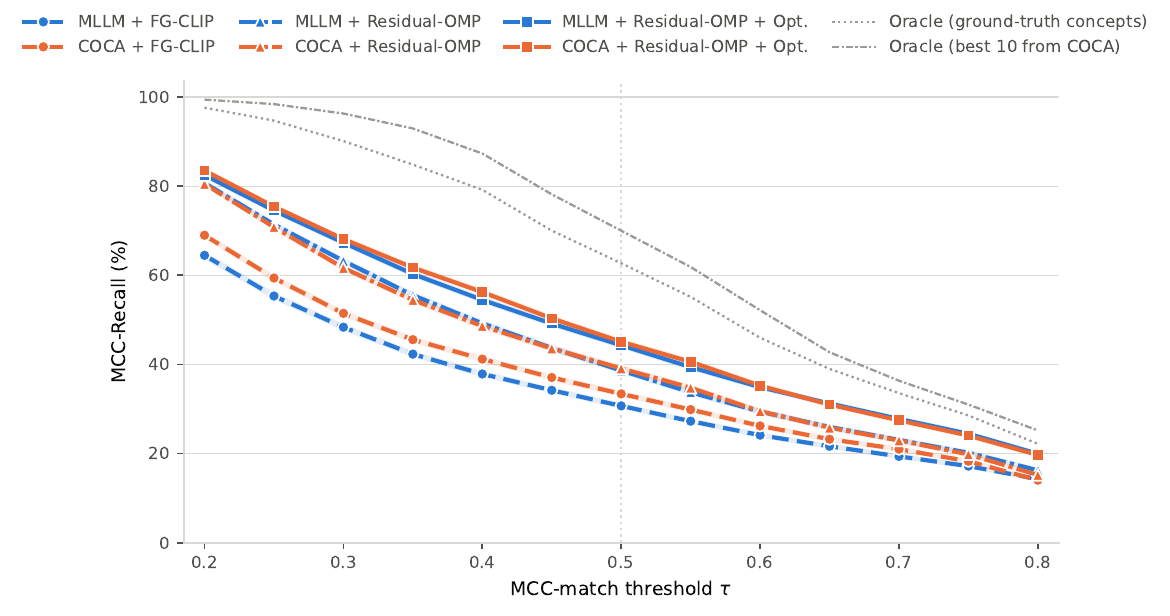}
\caption{Sensitivity analysis over the MCC-match threshold $\tau$. MCC-Recall (\%) across varying recovery thresholds $\tau \in [0.2, 0.8]$ for different concept discovery and scoring methods compared against oracle baselines. The vertical dotted line denotes the default threshold $\tau = 0.5$ used in the main experiments.}
\label{fig:supp:multi-concept:metrics:tau-sweep}
\end{figure}

One arbitrary design choice is the threshold $\tau = 0.5$ that indicates when we count a concept as recovered. Therefore, we ablate the effect of $\tau$ and sweep the range of $\tau \in \{0.2, 0.25, \ldots, 0.8\}$. Results are in \cref{fig:supp:multi-concept:metrics:tau-sweep}. We find that, as expected, MCC-R decreases with increasing $\tau$, as it becomes harder to confirm a concept. While the absolute value of the metric changes with $\tau$, the ordering of methods is stable up to the extremes, i.e., the choice of $\tau$ does not influence the order and relative gains of methods. The comparability of MLLM-proposed concepts and COCA concepts also holds true across thresholds.

\mypara{Embedding Space Cosine Similarity}
As in \cref{sec:evaluation} (\cref{sec:supp:ablations:judge}), we evaluate the concept recall by embedding both predicted and ground-truth concepts with the 
all-MiniLM-L6-v2 sentence embedding model. Then, for every ground-truth concept, we find the maximum cosine similarity across predicted concepts, and finally average the maximum similarities across concepts.

\mypara{LLM-Judged Recall}
We also measure recall by using an LLM to judge concept equivalence (metric $R_J$). In this case, we provide the list of ground-truth concepts and the list of predicted concepts to the LLM. The LLM then returns a list of all ground-truth concepts that have a related concept among the predicted concepts. Here, we define \enquote{related} as semantically equivalent, synonymous, or strongly associated. We also accept implied concepts, for example \enquote{stuffed animal} $\rightarrow$ \enquote{teddy bear}. The following (real) example illustrates how this metric is calculated:
\begin{center}
\begin{tabular}{l p{0.6\linewidth}}
Ground-Truth: & apple, backpack, car, cow, knife, \textbf{sheep}, \textbf{teddy bear} \\
\midrule
Predicted: & \textbf{toy bear}, \textbf{teddy bear}, wide open grassy field, pastoral scenery, livestock animal, rural setting, \textbf{sheep in a field}, \textbf{stuffed animal}, pasture field, \textbf{farm animal} \\
\midrule
Matched: & teddy bear, sheep
\end{tabular}
\end{center}
The LLM finds two (of 7) ground-truth concepts in the predicted concepts. Both ground truth concepts are matched with multiple predicted concepts each, and in the case of \enquote{sheep}, no predicted concept is identical to the ground-truth concept name. In this case, the recall $R_j = \frac{2}{7}$. 

\begin{figure}[h]
\centering
\includegraphics[width=\linewidth]{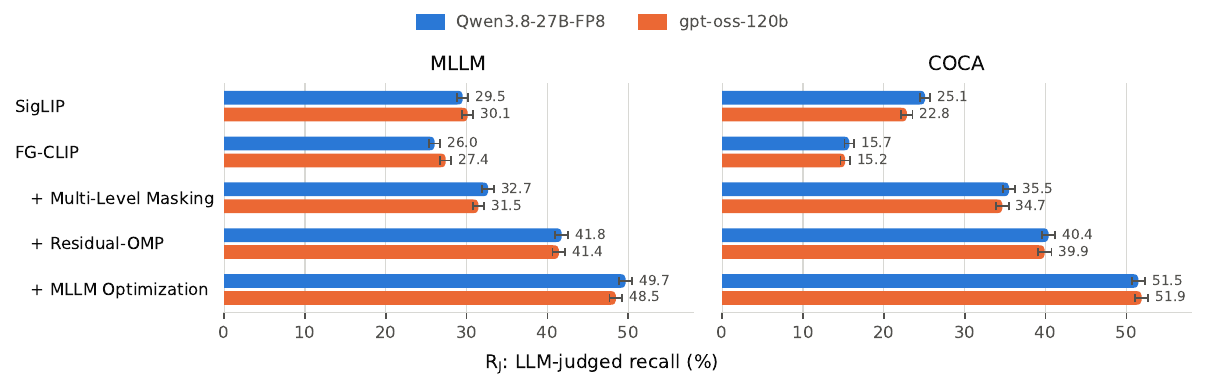}
\caption{LLM Judge Recall $R_J$ evaluated with Qwen3.8-27B and gpt-oss-120b. Scores are averaged over datasets. Error bars show 95\% CIs.}
\label{fig:supp:multi-concept:metrics:judgerecall}
\end{figure}

For the results in \cref{tab:complex_data_evaluation_recall}, we use Qwen3.8-27B as an LLM, but also calculate all metrics with gpt-oss-120B for comparison. The results are in \cref{fig:supp:multi-concept:metrics:judgerecall}. We find that scores differ slightly between Qwen3.8-27B and gpt-oss-120B, but not substantially, and they never flip the relative performance of any two methods, i.e., which one performs better.

Overall, we find that all three recall metrics largely agree with each other. Therefore, finally, we also analyze discriminative metrics on the setting in \cref{sec:multi-concept}. In \cref{sec:evaluation}, discrimination ($D$) is a useful metric because the target concept is present in almost all images, and all methods perform well. However, in the setting in \cref{sec:evaluation}, the prevalence of target concepts is much lower. In consequence, not even ground-truth concepts achieve meaningful separation between sets $A$ and $B$. Instead, we can only detect which concepts are more prevalent in set $A$ or in set $B$, but we cannot necessarily use them anymore to globally distinguish the sets.

To show this, we evaluate three alternative metrics on the same methods as in \cref{sec:multi-concept}, i.e., in \cref{tab:complex_data_evaluation_recall}. The alternative metrics are defined as follows:

\mypara{Set-Level Discrimination ($D$)}
This metric trains a logistic regression classifier on the continuous $f_{\mathrm{emb}}$ image-concept scores for the predicted concepts in 5-fold cross-validation. The classifier receives as input the 20 scores between an image and the concepts predicted for each set. From this, the classifier predicts the set membership of the image, i.e., $A$ or $B$.

\mypara{Concept-Level Discrimination ($D_c$)}
This metric works similarly to Set-Level Discrimination ($D$), but instead of predicting the set membership, the classifier predicts the presence or absence of ground-truth classes (or concepts) in the image. Specifically, for each set ($A$ and $B$), we train a logistic regression classifier in 5-fold cross-validation using only the continuous $f_{\mathrm{emb}}$ scores of the 10 concepts predicted for that specific set. The classifier's target is the binary ground-truth presence of each planted class. We evaluate the performance using the out-of-fold Area Under the Receiver Operating Characteristic (AUROC) rather than accuracy due to varying class prevalences, and average the scores over all ground-truth classes.

\mypara{Concept-Level Binarized Discrimination ($D_b$)}
This metric is identical to Concept-Level Discrimination ($D_c$), except that the continuous $f_{\mathrm{emb}}$ scores are binarized before training the classifier. For each predicted concept, we first find the ground-truth class it best matches and determine the threshold that maximizes the MCC for that match. The continuous scores for that concept are then binarized (0 or 1) using this specific threshold. We then train the same logistic regression classifier in 5-fold cross-validation on this binary matrix to predict the ground-truth class presences, again reporting the out-of-fold AUROC. While MCC-R evaluates if single concepts pass a strict threshold, $D_b$ measures whether the full set of predicted concepts, once reduced to binary indicators, can jointly decode the ground-truth classes.

Results are in \cref{tab:supp:multi-concept:discrimination}. We make the following observations: Discrimination ($D$) often yields scores that are only barely better than chance (50.0), including for ground-truth concepts. Only for DOTA, scores are better, reaching around 70\%. Furthermore, scores are inconsistent with the recall scores, which were generally lower for SUN RGB-D, but are higher for discrimination. We think the main reason for these problems is that the setting in \cref{sec:multi-concept} does not admit easy discrimination of sets because target concepts have overall low prevalence, and they are not entirely exclusive to either set. Another problem is fitting a classifier, which can overfit on spurious associations and use signals from all features combined instead of isolating individual concepts. Concept-Level discrimination metrics ($D_c$ and $D_b$) to some extent inherit this problem, but they agree better with recall metrics and generally show small variation between methods. Overall, we find using discrimination metrics difficult in this setting, and therefore focus on recall metrics.

\begin{table}[h]
\centering
\resizebox{\linewidth}{!}{
  \begin{tabular}{l@{\hspace{8pt}}cccc@{\hspace{8pt}}cccc@{\hspace{8pt}}cccc@{\hspace{8pt}}cccc}
  \toprule
  & \multicolumn{4}{c}{LVIS} & \multicolumn{4}{c}{MS COCO} & \multicolumn{4}{c}{SUN RGB-D} & \multicolumn{4}{c}{DOTA} \\
  \cmidrule(lr){2-5}\cmidrule(lr){6-9}\cmidrule(lr){10-13}\cmidrule(lr){14-17}
  {} & MCC-R & $D$ & $D_{\mathrm{c}}$ & $D_{\mathrm{b}}$ & MCC-R & $D$ & $D_{\mathrm{c}}$ & $D_{\mathrm{b}}$ & MCC-R & $D$ & $D_{\mathrm{c}}$ & $D_{\mathrm{b}}$ & MCC-R & $D$ & $D_{\mathrm{c}}$ & $D_{\mathrm{b}}$ \\
  \midrule
  RAM++ & 31.5 & \underline{57.8} & 72.8 & 72.6 & 44.7 & \underline{60.4} & 76.9 & 76.4 & 15.1 & 60.5 & 68.4 & 68.2 & 29.4 & 66.2 & 75.9 & 75.0 \\
  \midrule
  MLLM + SigLIP & 26.4 & 54.7 & 80.6 & 68.8 & 40.2 & 57.3 & 87.6 & 73.0 & 10.1 & 58.7 & 74.8 & 66.8 & 33.8 & 70.9 & \underline{87.4} & 74.6 \\
  MLLM + FG-CLIP, no masking & 22.2 & 56.5 & 78.6 & 70.3 & 27.4 & 58.8 & 84.1 & 73.4 & 15.8 & 61.5 & 72.8 & 66.5 & 31.1 & 69.3 & 83.3 & 74.1 \\
  MLLM + FG-CLIP & 27.5 & 54.9 & 80.5 & 72.2 & 42.4 & 57.5 & 87.8 & 77.6 & 20.6 & 60.8 & 75.2 & 69.9 & 32.4 & 68.8 & 83.3 & 74.3 \\
  \quad + Residual-OMP & \underline{32.8} & 56.2 & \textbf{82.9} & \textbf{78.1} & \textbf{53.5} & 58.9 & \textbf{90.7} & \textbf{84.5} & \textbf{22.9} & \underline{61.7} & \textbf{77.3} & \underline{73.2} & \underline{45.5} & 70.9 & 86.5 & \underline{82.4} \\
  \midrule
  COCA-20k + SigLIP & 24.3 & 54.9 & 79.5 & 69.2 & 40.7 & 57.4 & 87.1 & 74.8 & 6.6 & 59.0 & 75.3 & 67.9 & 34.9 & \underline{71.5} & \textbf{88.4} & 79.1 \\
  COCA-20k + FG-CLIP, no masking & 16.5 & \textbf{59.2} & 77.2 & 70.0 & 17.8 & \textbf{60.9} & 82.5 & 72.1 & 11.3 & \textbf{63.5} & 71.2 & 67.1 & 28.2 & \textbf{71.6} & 82.7 & 76.7 \\
  COCA-20k + FG-CLIP & 30.0 & 55.7 & 81.2 & 74.1 & 46.8 & 58.2 & 89.0 & 80.5 & 21.4 & 60.8 & 75.8 & 71.7 & 35.5 & 69.4 & 84.1 & 77.2 \\
  \quad + Residual-OMP & \textbf{33.4} & 56.7 & \underline{82.7} & \underline{77.6} & \underline{52.7} & 59.1 & \underline{90.5} & \underline{84.4} & \underline{22.6} & 61.4 & \underline{77.0} & \textbf{73.7} & \textbf{47.7} & \textbf{71.6} & 86.5 & \textbf{82.5} \\
  \midrule
  Oracle (ground-truth concepts) & 59.4 & 55.4 & 91.9 & 84.4 & 96.3 & 57.1 & 97.4 & 88.3 & 36.8 & 60.3 & 84.3 & 77.9 & 58.4 & 67.7 & 83.7 & 78.8 \\
  \bottomrule
  \end{tabular}
}
\caption{Comparison of MCC-Recall and alternative discrimination metrics across datasets. We report Set-Level Discrimination ($D$), Concept-Level Discrimination ($D_{\mathrm{c}}$), and Concept-Level Binarized Discrimination ($D_{\mathrm{b}}$) on LVIS, MS COCO, SUN RGB-D, and DOTA. Bold and underlined values denote the best and second-best performing methods.}
\label{tab:supp:multi-concept:discrimination}
\end{table}

\section{Technical Details of Applications}
\label{sec:supp:applications}

\subsection{Workflow for Finding Spurious Correlations in ImageNet}
\label{sec:supp:applications:imagenet}

To find spuriously correlated class-concept pairs in ImageNet, we proceed in three steps: (1) We determine a candidate pool of concepts where we search for spurious correlations with classes. (2) We filter the full product of classes $\times$ concepts using a number of tests, which involve finding surprisingly frequent associations, removing class-concept pairs where the class and concept are semantically closely related, and sorting candidates by likelihood and interestingness of potential correlations. (3) Finally, we verify a shortlist of 6,103 class-concept pairs with an MLLM.

\begin{figure}[t]
    \centering
    \includegraphics[width=\linewidth]{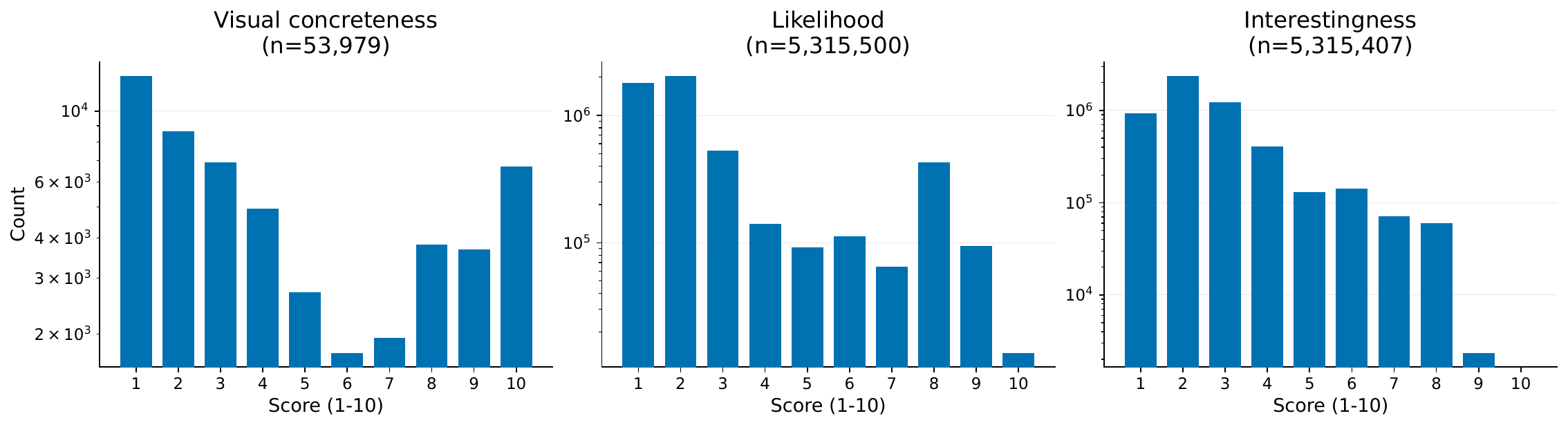}
    \caption{Distribution of LLM-assigned filtering scores. Histograms show the distributions of integer scores (1 to 10) across candidate concepts and class-concept pairs. Note the logarithmic count scale. \textit{Left:} Visual concreteness scores for concepts (Gemma4-31B). \textit{Center:} Plausibility scores for class-concept pairs (Qwen3.8-27B). \textit{Right:} Interestingness scores for class-concept pairs.}
    \label{fig:supp:applications:imagenet:scores}
\end{figure}

\mypara{Concept Candidates}
Our initial pool of candidates for spurious correlations is a subset of the COCA wordlist \citep{davies2010coca}. We take the 20,000 most frequent words, and for each word, we ask an LLM (Gemma4-31B in our case) how visually concrete the word is. The LLM assigns an integer score between 1 and 10 to each word, and we instruct it with the following examples:
\begin{alignat*}{3}
  &\text{\enquote{sunglasses}}    && \rightarrow 10 && \quad \text{(specific physical object)} \\
  &\text{\enquote{torn}}          && \rightarrow 9  && \quad \text{(directly visible material state)} \\
  &\text{\enquote{sponsored}}     && \rightarrow 6  && \quad \text{(partially concrete, grounded in visible logos)} \\
  &\text{\enquote{confrontation}} && \rightarrow 3  && \quad \text{(occasionally visually, e.g., two people facing off)} \\
  &\text{\enquote{predator}}      && \rightarrow 3   && \quad \text{(behavioral role, not visual feature)} \\
  &\text{\enquote{conservation}}  && \rightarrow 1  && \quad \text{(a policy/topic, not something a photo shows)}
\end{alignat*}
This instruction is deliberately biased toward detecting concrete class-object and class-attribute correlations, and not more abstract themes. We acknowledge that investigating this is of interest too, but abstract concepts are harder to verify, especially with automatic methods. After obtaining the scores from Gemma4-31B (see \cref{fig:supp:applications:imagenet:scores}, left panel, for their distribution), we keep all concepts with concreteness score $\geq 8$, which leaves 5,317 concepts.

\mypara{Filtering Class-Concept Pairs}
Next, we filter the full product of 1000 classes $\times$ 5,317 concepts. Concretely, we drop pairs where the embedding cosine similarity of class name and concept is $\geq 0.7$ using the EmbeddingGemma-300m model \citep{vera2025embeddinggemma}. Also, we calculate the FG-CLIP similarity scores $f_{\mathrm{emb}}(s, x)$ between all concepts and images. For each image, we only keep the scores of the 100 top ranking concepts by $z$-score, i.e., the 100 concepts with the highest score, and set all other scores to 0. Here, we first standardize concept similarity scores based on their per-concept mean and standard deviation across all images before comparing them. Based on this, we remove class-concept pairs where the concept has a nonzero score for $< 20$ images of the class.

Using the sparsified scores, we conduct two tests for each class-concept pair. First, we test whether the concept can separate the class from all other classes (one-vs-rest) with AUROC at least 0.55. In our formulation, AUROC is the rank-sum Mann-Whitney U statistic:
\begin{equation}\label{eq:supp:imagenet:auroc}
    \operatorname{AUROC}(c, s) = \frac{R_{c, s} - 0.5\cdot n_c (n_c+1)}{n_c \cdot (N - n_c)},
\end{equation}
$N$ is the total number of images in the target and background sets combined, $n_c$ is the number of images in class $c$, and $R_{c, s}$ is the summed rank of the images in class $c$ when sorting all ImageNet images by the concept similarity $f_{\mathrm{emb}}$. Note that here, higher scoring images have \textit{higher} rank. We then keep class-concept ($c$-$s$) pairs where $\operatorname{AUROC}(c, s) \geq 0.55$. Additionally, we keep a maximum of 200 concepts with the highest $\operatorname{AUROC}$ per class at this stage. Overall, this step finds class-concept pairs where the concept appears more frequently in images of the class than expected, using the entire ImageNet as the background distribution. Note that this is an instance of image dataset diffing.

Next, we filter class-concept pairs with two scores representing interestingness and plausibility of the correlation. The purpose of this stage is to create a shortlist of candidate pairs for expensive MLLM verification. Pairs that survive the previous filtering steps are either genuinely correlated in ImageNet, or noise introduced by FG-CLIP biases, such as high visual or semantic similarity although the concept is not actually present. The interestingness score targets concepts that are genuinely correlated, but not unexpected. For example, it is not surprising that sheep often appear on pastures, or cars in urban areas. Hence, these combinations should receive low interestingness scores. The plausibility score targets noise and FG-CLIP biases. For example, pairs like hamster-aghast or wild\_boar-dissolved are nonsensical and should be discarded. To get interestingness and plausibility scores, we ask an LLM (Qwen3.8-27B in our case) with class-concept pairs and appropriate elicitation prompts to provide integer scores between 1 and 10. See \cref{fig:supp:applications:imagenet:scores} (center and right panels) for their distribution. To improve computational efficiency, we generate scores for 10 concepts in each inference step at once. Class-concept pairs need a minimum likelihood score $\geq 3$ and minimum interestingness score $\geq 5$ to pass this stage.

We then compare surviving concepts to their distribution in the 20 most similar classes by concept distribution. Here, we represent each class by the 5,317-dimensional vector whose components represent the mean concept similarity scores $f_{\mathrm{emb}}$ over images in the class. Similarity between classes is determined by cosine similarity of these vectors. We include this step because we noticed that many trivial associations, such as dog-collar, are not filtered by the previous test. Here, \enquote{collar} is indeed more frequent in dog classes than in ImageNet in general, but it is not a special or unexpected correlation. In these cases, we find similar trends in neighboring classes. We calculate $\operatorname{AUROC}(c, s)$ in the same way as \cref{eq:supp:imagenet:auroc}, only changing the background corpus, i.e., replacing the entire ImageNet dataset with images from the 20 most similar classes. In this stage, the threshold a class-concept pair has to pass is $\operatorname{AUROC}(c, s) \geq 0.55$, and we only keep the 15 highest scoring concepts per class.

\mypara{MLLM Verification}
To verify the shortlisted class-concept pairs, we prompt an MLLM (Qwen3.8-27B) to determine whether the concept $s$ is clearly and visually present in images of the class $c$. Instead of exhaustively testing all images in a class, we use a batched verification process. For each candidate pair $(c, s)$, we rank the images in class $c$ in descending order by their FG-CLIP similarity score $f_{\mathrm{emb}}(s, x)$. We then query the MLLM in batches, i.e., we first test the $k$ highest-scoring images, then the next batch of $k$ images, and so on. We use $k=20$, but the first two tests are performed after 20 images, then 200 images, and then after each batch of 20 images. After each batch, we perform a binomial test (with Benjamini-Hochberg FDR control) to determine if the prevalence of the concept in the class is not lower than 10\%. If we find that with the given confidence level (we use $\alpha=0.05$) the candidate appears in less than 10\% of images, we drop it early to reduce computational cost. We return all concepts that survive all batches, i.e., are tested on all images of the class and appear in at least 10\% of them. In our case, this yields 1,553 class-concept pairs.

\subsection{Workflow for Finding Demographic Biases in Text-to-Image Models}
\label{sec:supp:applications:bias}

To systematically detect and verify demographic biases in text-to-image (T2I) models, we proceed in three steps: (1) We generate large-scale datasets of images using prompts with and without specific demographic descriptors to form contrast sets. (2) We extract and filter a candidate pool of biased concepts using vision-language models (FG-CLIP and SigLIP) coupled with Residual-OMP for diverse concept selection. (3) Finally, we verify the presence and statistical significance of these candidate biases using an independent MLLM.

\mypara{Image Generation}
We evaluate biases for country (10 categories), religion (5 categories), and race (7 categories) terms. For Flux 1-dev and Stable Diffusion 3.5 Large (as T2I models), we generate 100 images for every target prompt with template 
\begin{center}
\enquote{\texttt{a \{country$\mid$religion$\mid$race adjective\} \{man$\mid$woman\} \{is working as occupation$\mid$doing activity\}}}.  
\end{center}
Here, we use 60 occupations and 45 common activities. This means that, per demographic target, we generate 10,500 images. For each prompt, we also generate a corresponding baseline set of 10,500 images using the exact same prompt but omitting the demographic target, i.e., the country/religion/race specifier. This pairs a target distribution (e.g., \enquote{an Egyptian woman}) against a baseline distribution (e.g., \enquote{a woman}), which allows us to isolate visual concepts whose frequency is uniquely amplified or reduced by the country/religion/race term.

\mypara{Finding and Filtering Bias Concepts}
We reuse the candidate concept vocabulary from our ImageNet analysis (\cref{sec:supp:applications:imagenet}), i.e., 5,317 visually concrete concepts from the COCA wordlist. Using these concepts, we first compute the image-text similarity scores $f_{\mathrm{emb}}(s, x)$ for all concepts across all generated images using both FG-CLIP and SigLIP models.  To determine if a concept is disproportionately associated with the demographic term, we compare its score distributions in the target image set versus the baseline image set. We quantify the separation using the AUROC (\cref{eq:supp:imagenet:auroc}), i.e., the probability that a randomly chosen image from the target distribution (with demographic term specified) scores higher than a randomly chosen image from the baseline distribution (without demographic term). We conduct a two-tailed Mann-Whitney U test and apply Benjamini-Hochberg False Discovery Rate (FDR) control across all concepts to confirm candidates.

To ensure diversity in the final candidate list and reduce semantic redundancy, we select 30 concepts per prompt pair using the Multi-Level Masking and Residual-OMP methods proposed in \cref{sec:multi-concept}. Due to the wealth of discovered attributes and the cost of additional optimization, we do not use MLLM optimization in this application.

\mypara{MLLM Verification}
We check the candidates using an MLLM (Qwen3.8-27B). Instead of testing concepts on all generated images ($2 \times 10500$, computationally infeasible), we construct a stratified sample of 900 images per side (target and baseline prompt). We then ask the MLLM whether a concept is clearly visible in the image. To reduce inference cost, we query the MLLM with chunks of 10 concepts at a time. Finally, we perform the Fisher exact test on the MLLM's binary presence judgments to compare the prevalence of a concept in the target distribution against the baseline. Only candidates that show a statistically significant shift in visual presence that aligns with the direction predicted by the SigLIP or FG-CLIP embeddings are confirmed as genuine demographic biases. Additionally, we require that the prevalence in the target distribution differs from the baseline distribution by at least 10\% according to MLLM presence counts.